\documentclass[preprint, review, 12pt]{elsarticle}

\usepackage{amssymb}
\usepackage{amsmath}
\usepackage{subcaption}
\usepackage{booktabs}
\usepackage{array}

\usepackage{rotating}
\usepackage{babel}
\usepackage{url}
\PassOptionsToPackage{hyphens}{url}
\usepackage{acronym}
\usepackage{multirow}

\usepackage{tikz}
\usetikzlibrary{shapes.geometric, arrows} 
\tikzstyle{rect} = [rectangle, rounded corners, minimum width=3cm, minimum height=1cm, text width=3cm, text centered, draw=black] 
\tikzstyle{arrow} = [thick,->,>=stealth]

\begin{document}

\hyphenation{un-super-vised}
\hyphenation{semi-super-vised}
\hyphenation{re-quire-ments}
\hyphenation{clas-si-fi-ca-tion}

\newacro{CNN}[CNN]{convolutional neural network}
\newacro{IIC}[IIC]{invariant information clustering}
\newacro{LULC}[LULC]{land use and land cover}
\newacro{SVM}[SVM]{support vector machine}
\newacro{RF}[RF]{random forest}
\newacro{MLP}[MLP]{multi-layer perceptron}
\newacro{OBIA}[OBIA]{object-based image analysis}
\newacro{TTA}[TTA]{test-time augmentation}
\newacro{AP}[AP]{average precision}
\newacro{ROC-AUC}[ROC-AUC]{return-on-characteristics area under curve}
\newacro{NDVI}[NDVI]{Normalized Difference Vegetation Index}
\newacro{NIR}[NIR]{Near-infrared}
\newacro{ML}[ML]{Machine learning}
\begin{frontmatter}


\title{Habitat classification from satellite observations with sparse annotations}

 \author{Mikko Impiö\corref{cor1}\fnref{label1}}
 \ead{mikko.impio@syke.fi}
 \cortext[cor1]{Corresponding author}
\author{Pekka Härmä\fnref{label2}}
\author{Anna Tammilehto\fnref{label3}}
\author{Saku Anttila\fnref{label2}}
\author{Jenni Raitoharju\fnref{label1}}

 \fntext[label1]{Programme for Environmental Information, Finnish Environment Institute, Finland}
 \fntext[label2]{Data and Information Centre, Finnish Environment Institute, Finland}
  \fntext[label3]{Metsähallitus, National Parks, Finland}

\begin{abstract}
Remote sensing benefits habitat conservation by making monitoring of large areas easier compared to field surveying especially if the remote sensed data can be automatically analyzed. An important aspect of monitoring is classifying and mapping habitat types present in the monitored area. Automatic classification is a difficult task, as classes have fine-grained differences and their distributions are long-tailed and unbalanced. Usually training data used for automatic land cover classification relies on fully annotated segmentation maps, annotated from remote sensed imagery to a fairly high-level taxonomy, i.e., classes such as forest, farmland, or urban area. A challenge with automatic habitat classification is that reliable data annotation requires field-surveys. Therefore, full segmentation maps are expensive to produce, and training data is often sparse, point-like, and limited to areas accessible by foot. Methods for utilizing these limited data more efficiently are needed.

We address these problems by proposing a method for habitat classification and mapping, and apply this method to classify the entire northern Finnish Lapland area into Natura2000 classes. The method is characterized by using finely-grained, sparse, single-pixel annotations collected from the field, combined with large amounts of unannotated data to produce segmentation maps. Supervised, unsupervised and semi-supervised methods are compared, and the benefits of transfer learning from a larger out-of-domain dataset are demonstrated. We propose a \ac{CNN} biased towards center pixel classification ensembled with a random forest classifier, that produces higher quality classifications than the models themselves alone. We show that cropping augmentations, test-time augmentation and semi-supervised learning can help classification even further.
\end{abstract}



\begin{highlights}
\item We focus on the challenging problem of automatic fine-grained biotype classification in remote sensed imagery using sparse pixel-wise ground-truth annotations.
\item We propose a region-based approach, where CNNs are used to identify only the center-pixel using a wider region around it.
\item We compare unsupervised pretraining, supervised pretraining using a large out-of-domain dataset, and semi-supervised training as approaches to improve our CNN model.
\item We propose an ensemble of a \ac{RF} and our CNN approach with test-time augmentation and achieve the most promising results with this ensemble approach.
\item We make all our \ac{ML} implementations along with our field survey based annotations publicly available.
\end{highlights}

\begin{keyword}
habitat classification \sep machine learning \sep semi-supervised learning \sep transfer learning \sep random forests



\end{keyword}

\end{frontmatter}

\acresetall

\section{Introduction}
\label{sec:introduction}

Earth observation based land cover classification, where each pixel of a satellite raster is classified as belonging to a discrete class, is a well-studied problem for classes such as roads, farmlands, and urban areas \cite{turkoglu2021crop, wei2021scribble, huang2018urban}. As the number of classes increases, variability between classes decreases, making classification harder. \cite{mahdavi_remote_2018, van_horn_inaturalist_2018}. Finely-grained categories are found in commonly used habitat taxonomies, such as in the Natura2000 system defined in the EU Habitats Directive \cite{natura2013european}. Deep learning methods have become widely-used in land cover classification lately \cite{kentsch_computer_2020, mou2019learning, sherrah_fully_2016, nalepa2020unsupervised, turkoglu2021crop, marmanis2015deep}, but the size of available annotations remains a challenge for habitat classification.

Global biodiversity is declining due to human actions \cite{ipbes2019global, hoekstra2005biomecrisis, foley2005global, anderson2018biodiversity}. Habitat mapping is a fundamental component of biodiversity monitoring, making it possible to better understand biodiversity loss and the impacts of climate change \cite{turner2003remote}. In addition to the intrinsic value habitats have, they also provide important ecological services, such as production of oxygen and food, carbon capture and water filtration \cite{cardinale2012biodiversity, dasgupta2021economics}. Habitat conservation and mapping is a high priority in the EU Biodiversity Strategy for 2030 \cite{eubiodivstrategy}, aiming to enlarge existing Natura2000 conservation areas and provide protection for endangered habitat types. The EU Habitats Directive \cite{habitats_directive1992} also requires the classification and mapping of endangered habitats. However, production of up-to-date data on distribution and state of these endangered habitats is a difficult task, especially in remote areas.

On-site field surveys and aerial photographic remote sensing are the two main methods for habitat mapping \cite{lengyel2008habitat}. Field surveys provide high-quality, in-situ information, but conducting them is expensive especially in remote areas \cite{mumby1999cost}. Remote sensing can provide information on a much larger scale, but with reduced information content, as some information only visible in the field is always lost. However, the information gained via remote sensing can be close to the explanatory power of field surveys, making remote sensing a cost-effective approach \cite{rhodes2015relative}. Remote sensing is also the only approach that enables large-scale ecosystem and community observation, global coverage and continuous monitoring. \cite{reddy2021remote}.

The ability to discriminate different habitats from remote sensed imagery depends highly on the chosen class mapping \cite{corbane2015remote}, which is referred to as class taxonomy in our study. Some class taxonomies are based on differentiating habitats on-site, while others are designed especially for remote sensing purposes, focusing on large, spatially contiguous areas \cite{corbane2015remote}. This study focuses on the Natura2000 class system \cite{natura2013european}, which is an in-situ class taxonomy and challenging for remote sensing purposes.

Remote sensing is a particularly effective mapping method in remote and spatially extensive areas, such as in Northern Lapland, Finland, which is the focus area of our study. A number of endangered fell habitat types are endemic to Lapland \cite{kontula2019threatened, janssen2016european}, and reliable land cover mapping helps in their conservation. Field surveys have to be conducted during the short summer period and is a demanding and laborious task, as surveyors have to spend long periods of time in the field, far away from human settlements. Field surveys are limited due to difficult and time-consuming access to many remote areas, making automated remote sensing approaches necessary for covering the full variance of habitat types efficiently. While field surveys remain irreplaceable, as ground truth data is always needed for building models, remote sensing may also help targeting the costly field surveys to the most promising areas for improving models.

\Ac{LULC} classification of remote sensed grids has been studied extensively for natural environments, e.g., in \cite{van2014random, salovaara2005classification, sun2020improved}. Traditional machine learning based models, such as \acp{RF} \cite{magnusson_shrub_2021, gislason_random_2006}, \acp{SVM} \cite{liu_self-trained_2013}, \acp{MLP} \cite{mishra2016remote, yuan2020deep}, and maximum entropy models \cite{stenzel_remote_2014} have been applied to the problem. \Acp{RF} have proven to be especially useful in \emph{pixel-based} classification, where each pixel is classified separately, and have been used for classifying habitats \cite{mcdermid2005remote, van2014random}, croplands \cite{teluguntla201830} and tree species \cite{mayra_tree_2021}. However, pixel-based classification fails to leverage the useful information that the local environment may contain and some classes, such wetlands, cannot be distinguished from each other based on a single pixel \cite{mahdavi_remote_2018}. 
\emph{Object-based} classification, where objects formed by pixels of similar characteristics are classified based on features extracted from the whole object, is another commonly used approach for \ac{LULC} classification \cite{liu2010object, yuehong2018enhancing, bin2020object}. A critical preprocessing step for object-based classification is segmenting the image into objects of similar characteristics and in particular under-segmentation may lead to poor results \cite{liu2010object}. Some works relying on traditional machine learning methods combine pixel-based and object-based classification \cite{yuehong2018enhancing, bin2020object}. 

Recently, also \acp{CNN} have been used for different \ac{LULC} tasks \cite{mayra_tree_2021, dong_very_2020}. \acp{CNN} take as their input an image patch and they typically output either an image of the same size defining a class for each pixel (semantic segmentation) or a vector showing the probabilities of the patch to belong to each class. The latter can be used either for single-label classification, where the patch is assigned to the most probable class \cite{mayra_tree_2021, dong_very_2020, mou2019learning} or multi-label classification where multiple classes can be assigned for a single patch \cite{kentsch_computer_2020}. The patch classification has been used, for example, for 4-class tree classification of airborne hyperspectral data \cite{mayra_tree_2021}, different land cover classification tasks on airborne hyperspectral data \cite{mou2019learning}, forest classification into deciduous and evergreen types using drone data \cite{kentsch_computer_2020}, and 10-class general land cover classification of Very High Resolution satellite imagery \cite{dong_very_2020}. Patch classification approaches either require uniform patches or sacrifice the pixel level accuracy and focus only on the majority class(es) within the patches.

Semantic segmentation has been used for urban environments \cite{sherrah_fully_2016, huang2018urban}, crop mapping \cite{turkoglu2021crop}, and general \ac{LULC} into distinct classes \cite{laban2021sparse}. A bottleneck for semantic segmentation is that models need to be trained with dense semantic segmentation maps, where each pixel is annotated. Due to annotation requirements, most works focus on easily accessible areas such as urban and agricultural areas
\cite{huang2018urban, turkoglu2021crop}, use very generic classes that can be annotated without field surveys \cite{laban2021sparse}, or rely on airborne data with higher resolution leading to lower areal coverage \cite{mayra_tree_2021, kentsch_computer_2020}. Producing segmentation maps from sparse annotations has been studied (\cite{hua2021semantic, wei2021scribble}), but the approaches still rely on full segmentation maps in addition to sparse annotations. An alternative approach was proposed in \cite{laban2021sparse}, where sparse annotations of satellite imagery are used to create synthetic segmentation maps for training, overcoming the need for real fully-annotated segmentation maps.

Remote sensing has been used for habitat mapping \cite{mcdermid2005remote}, tree species and forest classification \cite{mayra_tree_2021, kentsch_computer_2020}, as well as wetland classification \cite{mahdavi_remote_2018, magnusson_shrub_2021}. Remote sensing has vast possibilities in biodiversity assessment and bioindicator extraction \cite{petrou_review_2011}. Niche classification taxonomies such as the Natura2000 have been less used in lieu of more high-level and easier classes. Random forests have been used for grassland \cite{stenzel_remote_2014} and wetland \cite{kopel2016application} Natura2000 classes, but we could not find studies considering a large number of very different Natura2000 classes simulatenously like it is done in our study. Lack of studies can be explained by the Natura2000 class taxonomy being highly difficult for remote sensing purposes due to some classes being distinguishable only in-situ, or being too small to be detected remotely.

Our study differs from earlier studies by considering a general classification model - a single model that classifies land cover to all possible Natura2000 types, not focusing on a single class group such as grasslands or wetlands. The large amounts of unannotated data for pretraining or semi-supervised learning has not been used for habitat classification previously. Our proposed methods solve problems in sparsely-annotated land cover classification, and is the first study where \ac{CNN}-based methods have been used for classification of arctic habitats via remote sensing.

\begin{figure}
        \centering
        \includegraphics[width=1\textwidth]{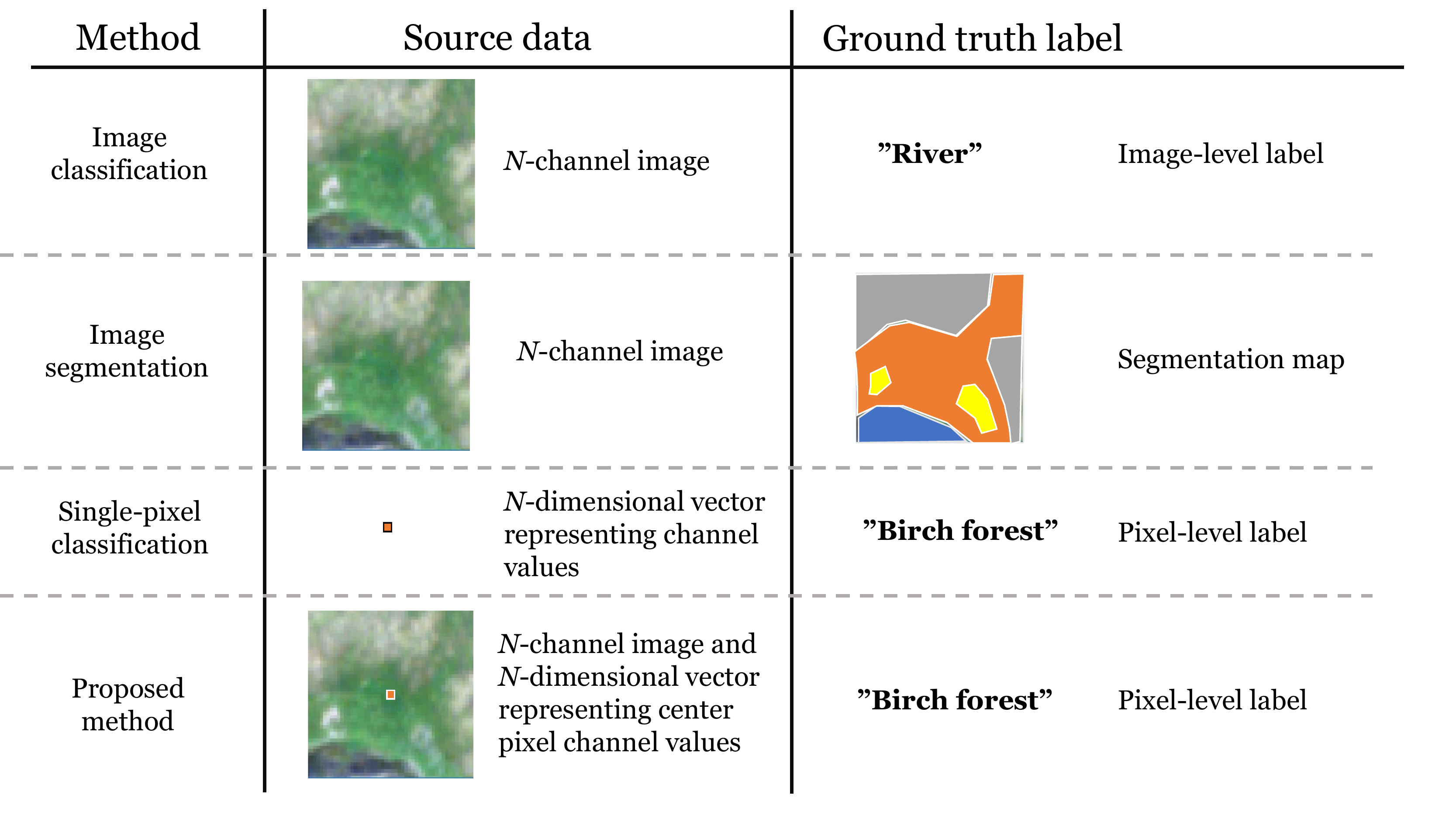}
        \caption{How the proposed method compares to other classification tasks}
        \label{fig:methods}
    \end{figure}

In this paper, we focus on the challenging problem of with fine-grained habitat classification from satellite and airborne measurements using very sparse pixel-wise annotations. We propose a region-based classification scheme, where the surrounding area is exploited in classifying the central pixel as illustrated in Fig.~\ref{fig:methods}. This approach is different from the object-based approach, because the aim is not to classify the full patch/object, but simply extract useful information from the surrounding area of the pixel of focus. Thus, there is no need for homogeneous objects/patches, while \acp{CNN} can be applied. To improve the \ac{CNN} classification with scarce training data we explore different state-of-the-art training techniques including transfer learning, unsupervised and semi-supervised training. Furthermore, we propose a novel approach for center-crop augmentation, a method for producing center-pixel biased CNNs. We also show the benefits of \ac{TTA} during inference, an approach that makes the final predictions more reliable.
We compare the performance, including the advantages and disadvantages, of the \ac{CNN}-based methods with \acp{RF} trained for pixel-wise classification. Finally, we achieve the best performance by combining the pixel-based \ac{RF} classifier with the region-based \ac{CNN}.

The main contributions of this paper are as follows:
\begin{itemize}
    \item We propose a region-based approach for pixel-based habitat classification where a \ac{CNN} and \ac{RF} ensemble is used to identify the center-pixel of a patch using a wider region around it.
    \item We compare different pretraining approaches with post-training semi-supervised learning.
    \item We show how beneficial it is to apply pretraining with out-of-domain data, for example using the CORINE land cover datasets.
    \item Our experiments show that the region-based ensemble model outperforms both \acp{RF} and \acp{CNN} separately.
    \item We also analyze the suitability of the methods to assist in locating certain habitats in field-surveys and to guide future field data collection.
    
\end{itemize}

The rest of this article is organized as follows. Section~\ref{sec:material} discusses the field-collected data and remote sensing imagery used for our experiments. Section~\ref{sec:methods} discusses the details of our proposed model and training approaches, as well as implementation details. Section~\ref{sec:results} displays the results and Sections \ref{sec:discussion} and \ref{sec:conclusions} discuss and conclude the results.
	
\section{Materials}
\label{sec:material}

\subsection{Annotations}

\subsubsection{Field survey-based annotations}

    \begin{figure}
        \centering
        \includegraphics[width=1\textwidth]{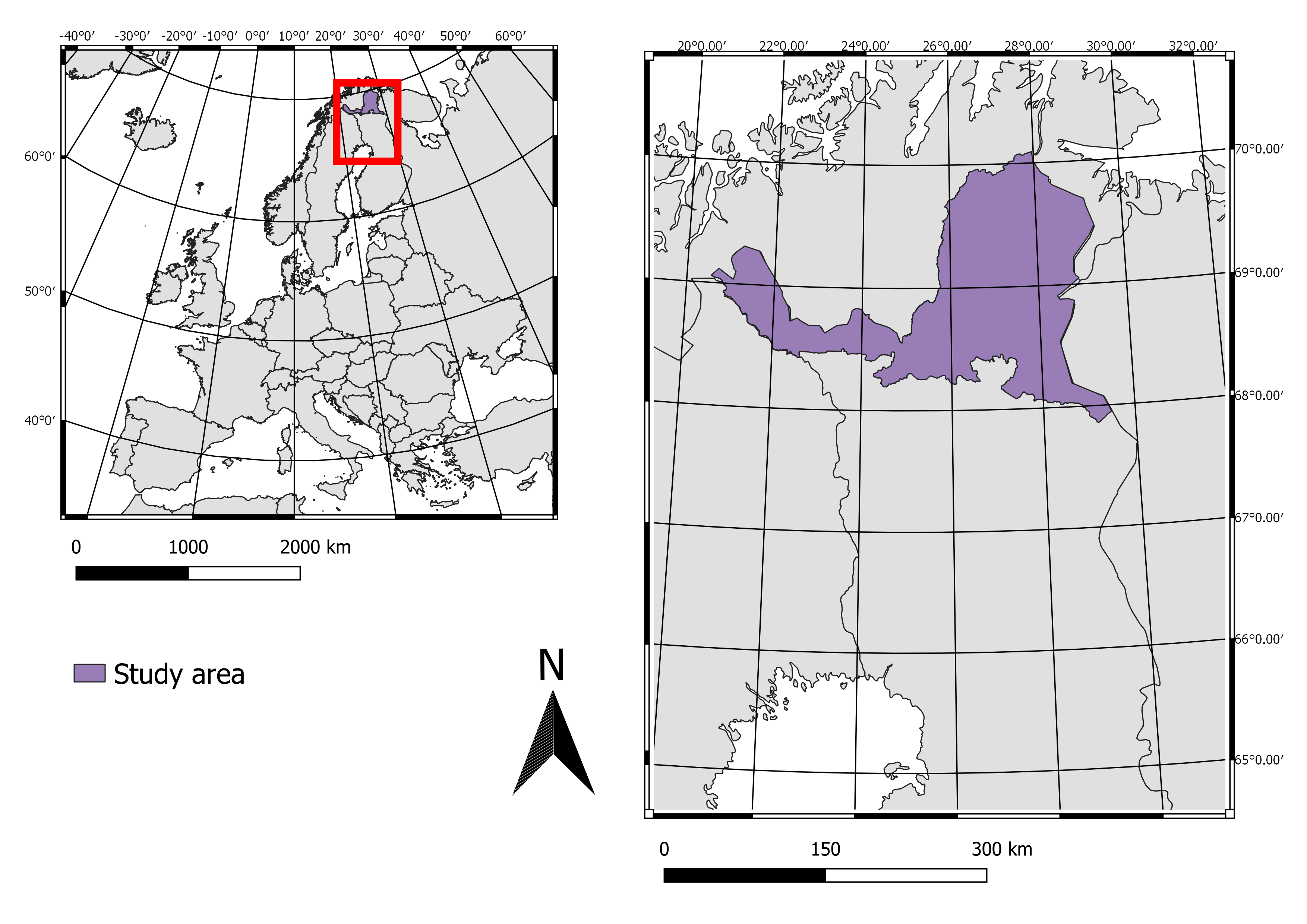}
        \caption{The study area in northern Lapland}
        \label{fig:lapland}
    \end{figure}

The classification ground-truth is based on field surveyed annotations. A total of 2558 georeferenced points representing the environment around a 10m radius was collected by Metsähallitus from Northern Lapland during summer 2020. Field data collection was done as a part of a project aiming at updating the habitat maps in nature conservation and wilderness areas in Northern Lapland. The data was collected mostly by foot by professional nature surveyors. The study area is shown in Figure \ref{fig:lapland}.

The collected dataset is rich in information and contains several attributes, for example habitat homogeneity and vegetation height. Two different class labels based on two different classification taxonomies, namely Natura2000 \cite{natura2013european} and the general classification system for Finland's biotopes \cite{tuominen2001yleispiirteinen}, are provided for most of the points.

Natura2000 is a commonly used classification taxonomy consisting of 233 different classes \cite{natura2013european}. The Natura2000 classification taxonomy is designed for nature conservation and focuses on endangered habitat types. The taxonomy is hierarchical in four levels, having an unique identifier for each class determining the hierarchy. For example the '7310 Aapa mires' class  belongs to the '73XX Boreal mires' group, which belongs to the highest '7XXX Raised bogs and mires and fens' group. The  dataset we are using has in total 25 unique Natura2000 classes, presented with their names and number of points in Table \ref{tab:naturanames}.

The collected dataset has some limitations. Some of the collected points do not have class information for both taxonomies, making the dataset size for the two taxonomies different. The total number of Natura2000-annotated points is 2036. The dataset is also highly imbalanced, as seen in Table \ref{tab:naturanames}, with the largest class containing 472 samples and the smallest classes only a single sample. Although the dataset is a large field survey dataset, it is fairly small for machine learning purposes. To address the limitations of a small dataset, we collected a larger out-of-domain dataset for pretraining.

\begin{table}[t]
\centering
\scriptsize
\begin{tabular}{p{0.06\linewidth}  p{0.8\linewidth} p{0.06\linewidth}}
\textbf{Code}         & \textbf{English name \cite{natura2013european}}                                                                                      & \textbf{\#} \\\hline
3110                  & Oligotrophic waters containing very few minerals of sandy plains (Littorelletalia uniflorae)               & 28    \\
3160                  & Natural dystrophic lakes and ponds                                                                         & 1     \\
3220                  & Alpine rivers and the herbaceous vegetation along their banks                                              & 4     \\
4060                  & Alpine and Boreal heaths                                                                                   & 472   \\
4080                  & Sub-Arctic Salix spp. scrub                                                                                & 9     \\
6150                  & Siliceous alpine and boreal grasslands                                                                     & 121   \\
6270                  & Fennoscandian lowland species-rich dry to mesic grasslands                                                 & 1     \\
6430                  & Hydrophilous tall herb fringe communities of plains and of the montane to alpine levels                    & 13    \\
6450                  & Northern boreal alluvial meadows                                                                           & 46    \\
7140                  & Transition mires and quaking bogs                                                                          & 165   \\
7160                  & Fennoscandian mineral-rich springs and   springfens                                                        & 104   \\
7220                  & Petrifying springs with tufa formation   (Cratoneurion)                                                    & 8     \\
7230                  & Alkaline fens                                                                                              & 26    \\
7240                  & Alpine pioneer formations of the   Caricion bicoloris-atrofuscae                                           & 2     \\
7310                  & Aapa mires                                                                                                 & 27    \\
7320                  & Palsa mires                                                                                                & 17    \\
8110                  & Siliceous scree of the montane to snow levels (Androsacetalia alpinae and Galeopsietalia ladani)           & 7     \\
8210                  & Calcareous rocky slopes with chasmophytic vegetation                                                       & 2     \\
8220                  & Siliceous rocky slopes with chasmophytic vegetation                                                        & 64    \\
9010                  & Western Taïga                                                                                              & 271   \\
9040                  & Nordic subalpine/subarctic forests with Betula pubescens ssp. czerepanovii                                 & 453   \\
9050                  & Fennoscandian herb-rich forests with Picea abies                                                           & 58    \\
9080                  & Fennoscandian deciduous swamp woods                                                                        & 12    \\
91D0                  & Bog woodland                                                                                               & 19    \\
91E0                  & Alluvial forests with Alnus glutinosa and Fraxinus excelsior (Alno-Padion, Alnion incanae, Salicion albae) & 106   \\\hline
                      &                                                                                                            & 2036
\end{tabular}
\caption{Natura2000 classes present in our dataset, accompanied with their Finnish and English names according to \cite{natura2013european}, and the number of samples in the full dataset}
\label{tab:naturanames}
\end{table}

\subsubsection{Supplementary annotations}
	
The larger out-of-domain dataset consists of 35 600 points sampled from the CORINE land cover raster from 2018 \cite{corinedataset}. The sampled points were selected randomly, with the requirement of a minimum of 500 meters between points to avoid overlapping patches. CORINE is a general land cover class taxonomy, containing 44 classes in five major groups. Unlike the Natura2000 classes, the classes are general land cover categories, such as agricultural areas, artificial surfaces, and forests. The classes are coarsely-grained and contain also human-built environments. A more coarse taxonomy here means that a single class, for example 'Natural grasslands' in CORINE, may correspond to several Natura2000 classes.
	
\subsection{Remote sensing data}

High quality source data is important in land cover classification. Remote sensing data should cover a wide spectral and temporal range of observation with high spatial resolution. We use three different data sources: laser scanning, Sentinel-2 (S2) imagery, and \ac{NDVI} time-series aggregates derived from the S2 imagery.

The laser scanning data was collected from fly-over scannings performed in 2018-2020. Point clouds were processed into surface models with spatial resolution of 8m, including information on vegetation height and density. Sentinel imagery is mosaiced from imagery collected in the summer of 2020. A subset of 9 bands is used, leaving out the 60m resolution bands and the 8A band correlating highly with band 8. Sentinel 2 \ac{NIR} and red channels were used to calculate aggregated \ac{NDVI} rasters for the year 2020. \ac{NDVI} is a good indication of vegetation, and time-series of \ac{NDVI} data can be processed to produce rasters describing characteristics of the growth season. We use three rasters: \ac{NDVI} amplitude, \ac{NDVI} sum, and maximum \ac{NDVI}. Thus, our data consists of a total of 14 different channels.

The original data was produced by Finnish Environment Institute for other purposes. For this study, we processed the raw rasters further, resampling the rasters to uniform 10m resolution and calculating the mean and standard deviations across the full raster for standardization.

We used both pixel-based classification methods and methods taking the local neighborhood into account. Therefore, in addition to collecting pixel-wise training data for each of our annotated points, we collected a separate dataset for 49x49 pixel patches surrounding the annotated points.

\subsection{Train-test splits}

Preventing data leakage is challenging with georeferenced data. For \acp{CNN} training, we use patches collected around the ground-truth annotation points. A traditional random train-test split can not be used, as the patches that are geographically very close to each other, resulting in overlapping patches. The \ac{CNN} might learn some features present in the training set and validation on patches that contain the same features would not reflect the predictive power of the model. 

    \begin{figure}
        \centering
        \includegraphics[width=1\textwidth]{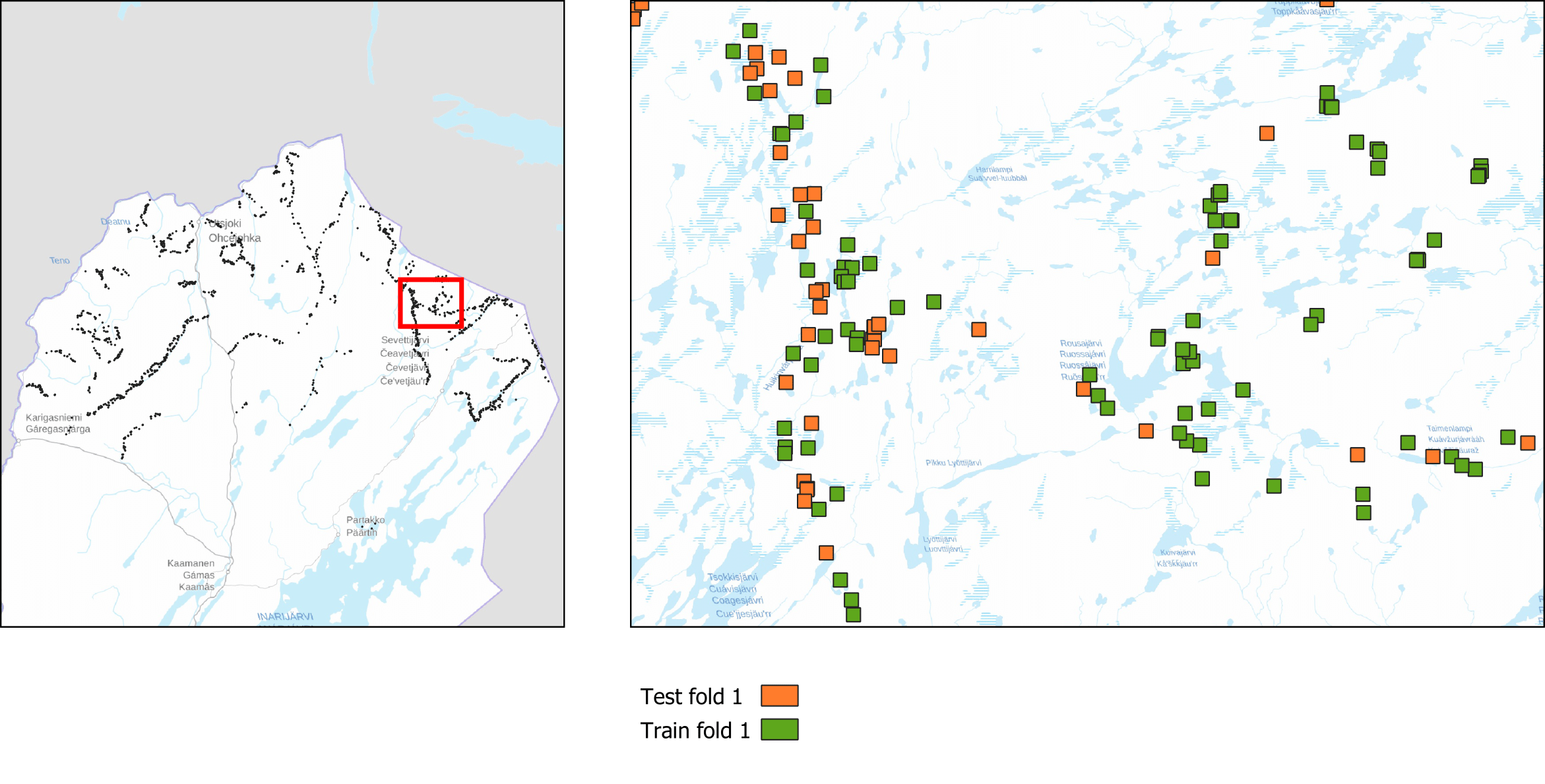}
        \caption{The field-collected data point locations}
        \label{fig:datapoints}
    \end{figure}

To address this problem, we chose a random test set of approximately 20\% of the original dataset and removed all the remaining patches that overlap with this set. The resulting non-overlapping patches then became the training set. The train-test splitting for the first fold of total 5 cross-validation folds is visualized in Figure \ref{fig:datapoints}.

This approach reduces data leakage, but has still some problems, as train and test patches might be collected near a common geographical feature, such as a river. A better approach would be using totally different regions for training and testing. However, as some habitat classes are very local, regional splitting would lead to problems in class balances, as all classes would not be present in all regions. 

\section{Methods}
\label{sec:methods}

\section{Overview of the applied methods}

To solve the problem of sparsely-annotated habitat classification from remotely-sensed data we compared many different approaches, including traditional pixel-based training, state-of-the-art deep learning algorithms, and some novel techniques proposed below. As our classifiers, we used commonly used \ac{RF} that classifies each pixel separately. We also formulated a \ac{CNN}-classifier which exploits also the surrounding of the pixel in its decision-making, but still only classifies the center pixel as illustrated in \ref{fig:methods}. Our final best-performing model is an ensemble of these pixel-based and region-based classifiers. The details of the classifier are described in Section~\ref{ssec:classifiers}.

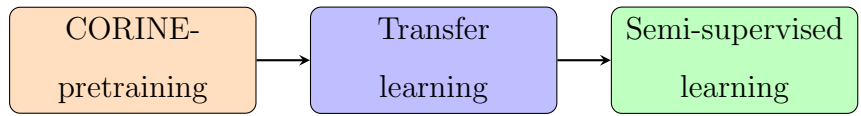
\begin{figure}
    \centering
    \begin{tikzpicture}[node distance=4cm]
        \node (pretrain) [rect, fill=orange!25] {CORINE-pretraining};
        \node (transfer) [rect, right of=pretrain, fill=blue!25] {Transfer learning};
        \node (semisup) [rect, right of=transfer, fill=green!25] {Semi-supervised learning};
        
        \draw [arrow] (pretrain) -- (transfer);
        \draw [arrow] (transfer) -- (semisup);
    \end{tikzpicture}
    \caption{Training phases}
    \label{fig:training}
\end{figure}

The models were trained in three (optional) phases, as shown in Figure \ref{fig:training}. In the first phase, we  pretrained the \acp{CNN} either in an unsupervised manner or using out-of-domain annotations. In the second phase, we applied transfer learning and trained the \acp{CNN} using our sparse annotations. In the third phase, we tested whether any performance improvement is gained using semi-supervised learning using a larger, unlabeled dataset. The details of the training approaches are given in Section~\ref{ssec:training}.

Data augmentation \cite{shorten2019survey} is another approach commonly used to improve \ac{CNN} training. While there are many standard tricks for data augmentation, many of them cannot be directly applied in our case due to the input channels representing different modalities and due to center-pixel focused classification approach. Therefore, we present a novel data augmentation technique based on random cropping of the surrounding area, along with test time data augmentation. Data augmentation details are discussed in Section~\ref{ssec:augmentation}.

\subsection{Classifiers}
\label{ssec:classifiers}

\subsubsection{\Aclp{RF}}

\acp{RF} have been frequently used in classifying remotely-sensed data \cite{gislason_random_2006, magnusson_shrub_2021}. They take as an input a vector containing information about a single pixel/polygon, which typically means different channels of the satellite data or some derived data products for that pixel/polygon, and they output a class label for that pixel (illustrated as single-pixel classification in Fig.~\ref{fig:methods}).

The \acp{RF} \cite{breiman2001random} is an ensemble of decision trees that are trained with different subsets of the original dataset, commonly referred to as bagging, as well as using different subsets of the original features. The main benefit of \acp{RF} are that they need significantly less data than \acp{CNN} or other deep learning models and need a lot less fine-tuning \cite{murphy2012machine}. \Acp{RF} also excel with unbalanced data \cite{khalilia2011predicting, zhou2012loan}. A drawback of \acp{RF} is that classification is pixel-based, which can be problematic because the semantic content of a pixel can change depending on the surroundings. For example, a palsa mire and an aapa mire are differentiated by occurence of permafrost mounds - features that are larger than a single pixel. Wetland areas between permafrost mounds are highly similar, and without contextual information, the in-between areas of palsa mires are classified as aapa mires.

\subsubsection{Region-based center-pixel focused \acl{CNN}}

\ac{CNN} classifiers differ from \acp{RF} and other traditional classifiers so that they take the whole image as their input. The most typical \ac{CNN} classifiers \cite{he2016deep, khosla_supervised_2020} output a class label for the whole image (illustrated as image classification in \ref{fig:methods}). Another typical approach is to use fully convolutional networks \cite{maggiori2016fullyconnected} that output an image that has equal size to the input images and contains class labels for each pixel (illustrated as image segmentation in \ref{fig:methods}). As we only know ground-truth for single pixels, but we assume that the surrounding pixels may contain valuable information for the decision-making, we formulate an approach, where the input is still an image, put the output is a class-label for the center pixel (illustrated as the proposed classification method in \ref{fig:methods}). 

While \acp{CNN} have outperformed traditional classifiers in several computer vision tasks \cite{yoo2019comparison, he2016deep}, their main drawback is the demand for very large amounts of training samples, which may be a bottleneck for applications such as ours, where annotating samples is slow and laborious. Nevertheless, also several techniques to overcome this drawback have been proposed in the literature \cite{xie2020self, ji2019invariant}, some of which are also considered in this work and discussed in the following sections.

\subsubsection{Ensemble classifier}

    \begin{figure}
        \centering
        \includegraphics[width=1\textwidth]{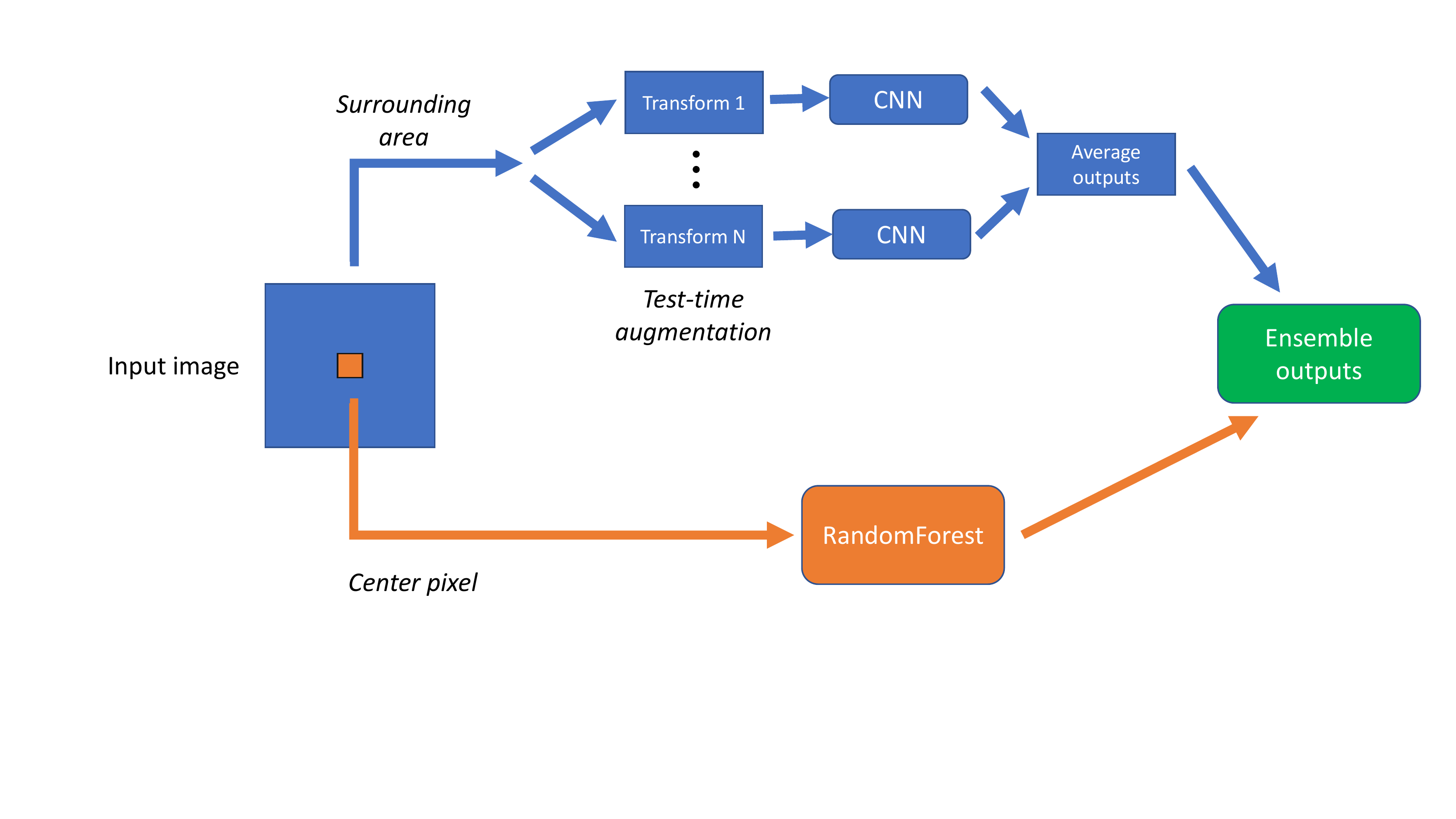}
        \caption{Classifying an image with the ensemble classifier}
        \label{fig:ensemble}
    \end{figure}
    
We introduce a novel approach combining the best sides of both \acp{RF} and \acp{CNN}. The model is based on training a \ac{CNN} and a \ac{RF} separately and using them as a stacked ensemble during inference. The ensemble addresses the drawbacks of both models: context-blindness of \acp{RF} and the large data requirements of \acp{CNN}.

Figure \ref{fig:ensemble} illustrates the classification process with the ensemble model. A center pixel $x$ and its surrounding patch $I$ are fed to the \ac{RF} and \ac{CNN} models, respectively, to produce outputs $y_{RF} = RF(x)$ and $y_{CNN} = CNN(I)$. When using \ac{TTA}, the image $I$ can be augmented before feeding it to the \ac{CNN}. Ensemble model combines the outputs of both the \ac{RF} and the \ac{CNN} to a final output vector $y$ by averaging the output separately for each class:
\begin{equation}
    y^i = \alpha y_{RF}^i + (1-\alpha) y_{CNN}^i,
\end{equation}
where $i$ is the class index and $\alpha$ is the weighting coefficient for classifier importance. We use an equal weight of $\alpha=0.5$ for all experiments.

\subsection{Training approaches}
\label{ssec:training}

We apply several different training approaches for center-pixel focused \ac{CNN} training. Pretraining and transfer learning are widely-used approaches and their benefits compared to random initialization of neural networks are well-known \cite{pan2009survey, weiss2016survey}. A large out-of-domain dataset can be used in pretraining, which has been shown to yield good results in several applications \cite{zhuang2020comprehensive}. If a large annotated dataset is not available, unsupervised pretraining can also help initializing the network weights for further fine-tuning with a smaller domain dataset \cite{erhan2010does, manas2021seasonal}. 

We apply the following three methods for \ac{CNN} training: unsupervised pretraining using \ac{IIC} approach by Ji et al. \cite{ji2019invariant}, out-of-domain pretraining with the CORINE dataset, and semi-supervised training using the Noisy student-method proposed by Xie et al. \cite{xie2020self}.

\subsubsection{Out-of-domain pretraining}

Pretraining classification models with out-of-domain datasets is a standard practice in deep learning, and has been shown to improve model performance \cite{erhan2010does, zhuang2020comprehensive}. A large out-of-domain dataset with the same number of input channels as the smaller target dataset is first used for initial training of the selected model and the model is later fine-tuned via transfer learning by training the final classification layer with the target dataset. The main benefit of this approach is that it provides the model with enough data to learn to extract some useful generic features, which are often useful also for the target data even if the classes are quite different. Even in earth observation, pretraining with a general purpose dataset containing all sorts of natural images, can improve transfer learning to the domain of \ac{LULC} classification \cite{marmanis2015deep}. We show that sampling points from the CORINE land cover inventory and using them to pretrain the final fine-grained model improves classification performance.

Arguably, the most common out-of-domain dataset for many specialized applications is the ImageNet dataset \cite{deng2009imagenet}, as ImageNet-pretrained weights are readily available for many model architectures. ImageNet pretraining can be used to train intermediate layers in the network, but the first convolutional layer typically needs to be trained as remote sensed imagery usually contains more channels than the three present in RGB images.
In our experiments, we show that a custom dataset, i.e., the CORINE land cover data \cite{corine} described in Section \ref{sec:material} can be successfully used in pretraining for biotype classification.
    
\subsubsection{Unsupervised pretraining using invariant information clustering}


\Ac{IIC}  is a image clustering method proposed by Ji et al. \cite{ji2019invariant}. It is based on applying a transformation $\Phi$ to image $x$ and finding a network configuration $Phi$ that maximizes the mutual information $I(\cdot,\cdot)$ between the output vectors of a neural network for both the original ($x$) and a transformed image ($x'$):
\begin{equation}
    \underset{\Phi}{\text{max}} \ I(\Phi(x), \Phi(x')).
    \label{eq:iic}
\end{equation}
In addition to minimizing the entropy between an original image and its transformation, this approach maximizes the intra-class entropy, addressing the problem of class-collapse, where all samples are clustered to only a few classes. The only hyperparameter that needs to be chosen is the number of clusters $C$. As the model is a \ac{CNN}, complex textures can be learned.

\subsection{Transfer learning}

Transfer learning refers to a practice where a model is initially trained with a different, often larger dataset than the final in-domain dataset. This often yields better results than starting the in-domain training with a random initialization of network weights \cite{weiss2016survey}. Initial training is often called pretraining, and datasets like the ImageNet are often used for the task. In this paper, we test both ImageNet, and a custom dataset collected from the data domain as the pretraining dataset.

The pretrained model is used as a starting point for the actual model training with data from the final classification domain. The final fully connected layer of the model is changed to match the number of classes in the final taxonomy and the model is fine-tuned using the final field-collected, ~3000 patch dataset with the desired taxonomy. In the transfer learning phase, either only the classification head or also the convolutional layers are trained, depending on the regularization methods and augmentations.

The practice where convolutional layer weights are not modified during training is referred to as freezing. Only the fully connected classification head weights are learned during the transfer learning phase. The frozen weights are used straight from the pre-trained model.

\subsubsection{Semi-supervised Noisy Student training}

Training with a small dataset can lead to overfitting the model to the dataset. Generality of the model suffers and it may not work on areas not present in the original training patch set. To combat this, we attempt to produce a model with better generalization by continuing the training using a large unlabeled dataset. The dataset patches are the same as the CORINE-annotated dataset, but it is now used without the CORINE class labels. We show that using semi-supervised learning with an unannotated dataset generally improves classifier performance in habitat classification.

Fully unsupervised and semi-supervised training, where a large pool of unannotated images are used during training has leaped forward in recent years, with several studies showing improvements in general purpose image classification \cite{grill_bootstrap_2020, chen_simple_2020, khosla_supervised_2020}. OUr semi-supervised training approach is based on Noisy Student training proposed by Xie et al. in \cite{xie2020self}. The method is based on knowledge distillation, pseudo-labeling and augmentation invariance. The finetuned model produced in the transfer learning phase is used as an teacher model to a new model that is trained from scratch. The teacher model produces pseudo-labels for the unlabeled patches in the large dataset, which are then used to train the student model. The student model is fed images from the small labeled dataset, with additional noise for regularization. Cross-entropy between the student model output and the soft pseudo-labels is minimized to find the parameters for the final model.

\subsection{Augmentations}
\label{ssec:augmentation}

\subsubsection{Training-time augmentation}
Augmentations plays a large role in successfully training a classifier. Image data augmentation by different geometrical and color transformations has been shown to improve generalization of a classifier and avoid overfitting \cite{shorten2019survey}. Although several augmentation methods for RGB images exist, the multispectral data presented as an N-channel image can not be augmented using the same techniques. The channels can contain data from different modalities and value ranges, where value changes can change the represented scene drastically. In our case, the annotated pixel is located in the center of an image and, therefore, geometric augmentations, such as cropping, need to take this into account. 
    
Due to the data and annotation constraints, we select only basic augmentation methods. Horizontal and vertical flipping does not affect the represented scene and the annotated pixel is unmoved. We experimented with value transforming augmentations and included random Gaussian blur as an additional augmentation. Gaussian blur is a suitable augmentation for remote sensing data because the value ranges can be arbitrary and the blurred scene is still a valid representation of the underlying true natural scene.
    
The center pixel annotation gives a possibility of using this information during augmentation. We propose using random center cropping as an additional augmentation method for images where the center pixel is annotated. Because the ResNet implementation uses an adaptive average pooling layer the input size can be changed between batches. During training, a batch is randomly cropped to a square with an odd side length between $3$ pixels and the maximum size of a training square, in this case $19$.

\subsubsection{Test-time augmentation}

Test-time augmentation can improve the robustness of a prediction by classifying a input several times with different augmentations applied. For the test-time-augmented results, we augmented and ran inference on each input region separately five times, and produced a final prediction based on the average of the softmax outputs of the network.

\section{Results}
\label{sec:results}

\subsection{Evaluation metrics}

The results are evaluated using commonly used classification metrics: Top 1,3, and 5 accuracies, f1-score, precision and recall. In addition, threshold metrics \ac{AP} and \ac{ROC-AUC} are also calculated. 

With imbalanced data, the choice of the averaging method across classes can have a large effect on the interpretation of the results \cite{forman2010apples}. We use the weighted mean across classes for each metric, calculating a metric for each class separately and weighing it depending on the amount of samples in the class. Other approaches, such as micro- or macro averaging are less useful in single-score metrics, as micro-averaging produces the same score for all metrics and macro-averaging (i.e., uniform weighted average) is biased towards the smallest classes. However, non-weighted micro- and macro averages are displayed for precision-recall curves in Figure \ref{fig:classwise_prec_rec}. Similar averaging needs to be done across cross-validation folds, where we use the common macro-averaging approach for calculating the final classifier performance metric and its standard deviation.

\subsection{Experimental setup}

\begin{figure}[!htb]
    \centering
    \includegraphics[width=1\textwidth]{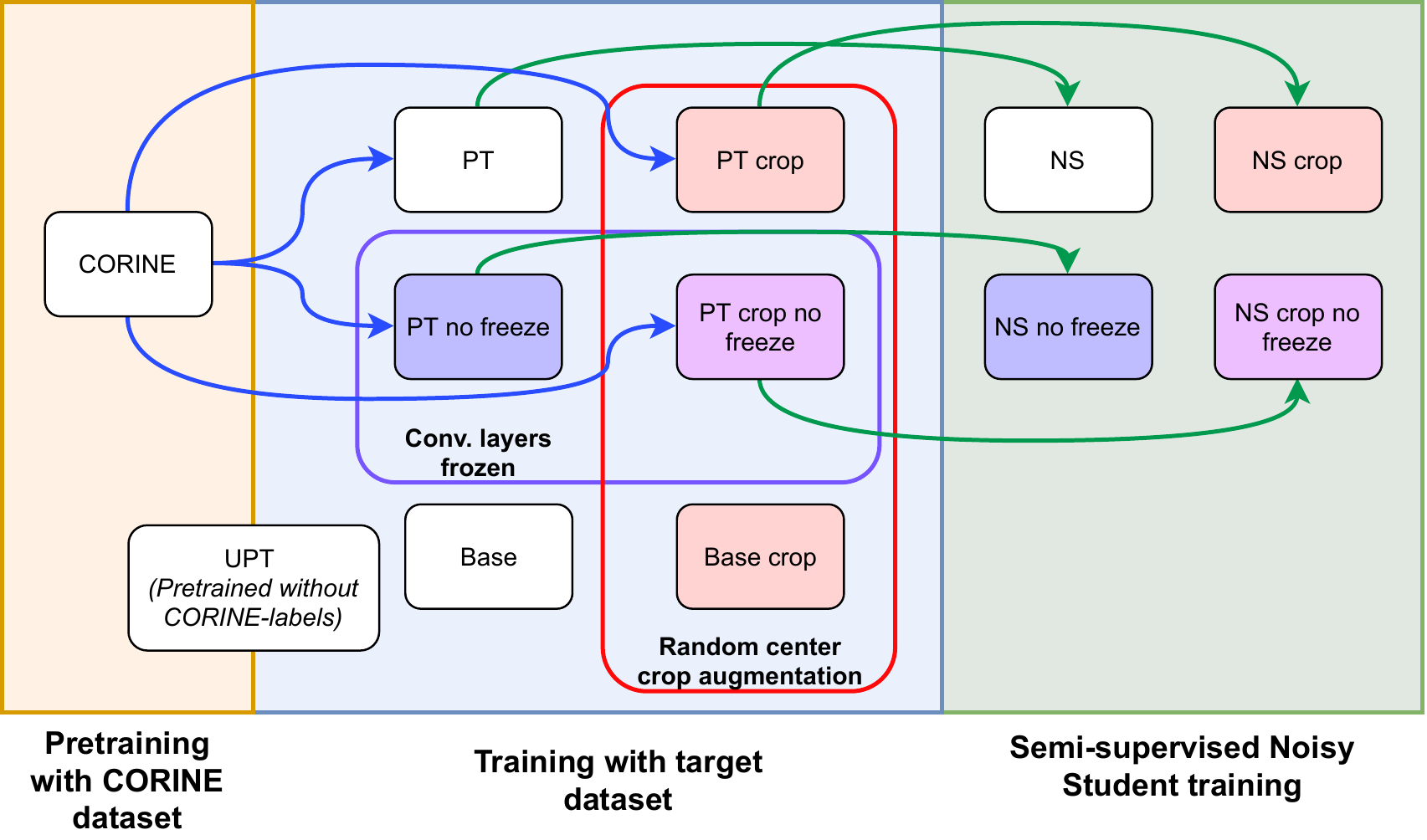}
    \caption{Relationships between different trained models. The Base models are trained only on the small target dataset. PT models are transfer learned from a model trained on a large CORINE-labeled dataset. NS models are a continuation of the PT models, using the PT models as a teacher model during training. The UPT model is pretrained unsupervised using the CORINE dataset without labels, and fine-tuned using the target dataset}
    \label{fig:training_phases}
\end{figure}

All the models trained in the three training phases seen in Figure \ref{fig:training_phases}. The three phases of pretraining, transfer learning and semi-supervised learning include 11 models trained with different approaches discussed in detail in Section \ref{sec:methods}. We trained different versions of models in the same phase, with different attributes applied. For example, the transfer learned models were trained with convolutional layers both frozen and trainable, as well as with random center cropping augmentation applied and not applied.

The transfer learned models were further trained with semi-supervised learning, resulting in different models with convolutional freezing and center cropping both applied and not.

We used a ResNet18 architecture \cite{he2016deep} as a backbone to all the \ac{CNN} models.  The first convolutional layer was changed to match the amount of channels in the image, and the final fully connected layer was changed to match the number of classes. All of the models were trained with same hyperparameters and optimizers to ensure comparability. Optimization was done using the Adam algorithm \cite{kingma2014adam} with a learning rate of 1e-4. Batch size of 128 was chosen for all tests. Series of test were conducted to choose training time of 500 epochs for the transfer learning and distillation tasks, with pretraining of 50 epochs for the CORINE- and \ac{IIC}-pretrained models.

For the \ac{RF} classifier that we used both independently and as a part of the ensemble classifier, we used a decision forest consisting of 100 decision trees trained with Gini impurity splitting. Hyperparameter optimization with grid and random search was done for different forest sizes, maximum tree depths and node impurity measures. Different parameters or increasing the amount of trees had little to no effect on the results, thus the most simple configuration of 100 trees and Gini impurity was chosen as the final model.

\clearpage

\begin{table}[ht]
\caption{Trained models and their attributes}
\centering
\begin{tabular}{l>{\centering}p{0.5cm}>{\centering}p{0.5cm}>{\centering}p{1.5cm}>{\centering}p{1.5cm}>{\centering}p{1.5cm}}
\toprule
 & \multicolumn{5}{c}{\textbf{Attribute}}\tabularnewline
\cline{2-6} \tabularnewline
& \multicolumn{2}{c}{\textbf{Pretraining}}\tabularnewline
\cline{2-3} \tabularnewline
\textbf{Model} & \begin{turn}{90}
Unsupervised
\end{turn} & \begin{turn}{90}
CORINE
\end{turn} & \begin{turn}{90}
{\tabular{@{}l@{}}Conv. layer \\freeze\endtabular} 
\end{turn} & \begin{turn}{90}
{\tabular{@{}l@{}}Random crop \\augmentation\endtabular} 
\end{turn} & \begin{turn}{90}
{\tabular{@{}l@{}}Semi-supervised \\learning\endtabular} 
\end{turn}\tabularnewline
\hline
\hline
Base              &           --                  & --  & --  & --  & -- \tabularnewline
Base crop         &           --                  & --  & --  & +   & --  \tabularnewline
\midrule
NS                &           --                  & +   & +   & --  & + \tabularnewline
NS crop           &           --                  & +   & +   & +   & + \tabularnewline
\multicolumn{1}{p{2.0cm}}{NS crop no freeze} & -- & +   & --  & +   & + \tabularnewline
NS no freeze      &           --                  & +   & --  & --  & + \tabularnewline
\midrule
PT                &           --                  & +   & +   & --  & -- \tabularnewline
PT crop           &           --                  & +   & +   & +   & -- \tabularnewline
\multicolumn{1}{p{2.0cm}}{PT crop no freeze} & -- & +   & --  & +   & -- \tabularnewline
PT no freeze      &           --                  & +   & --  & --  & -- \tabularnewline
\midrule
UPT               &           +                   & --  & +   & --  & -- \tabularnewline
\bottomrule
\end{tabular}

\label{tab:modelinfo}
\end{table}

\subsection{Evaluation results}

Table \ref{tab:nat_avg_table} shows the overall performance of different ResNet-models, the \ac{RF} baseline and the combined ensemble models. Model was evaluated with five-fold cross-validation. All evaluations were performed with \ac{TTA} with five augmentation rounds.

\begin{table}[htb]
\caption{\textbf{Ensemble results for Natura2000 classes:} The CNN models ensembled with the random forest model. Trained models include a baseline model (Base) trained only with the final small dataset, a CORINE-pretrained model (PT), model pretrained in an unsupervised manner (UPT), and the PT model continued with semi-supervised Noisy Student learning (NS). Some models have alternative models with cropping or convolutional layer freezing applied.}
\centering
\begin{tabular}{lcccc}
\toprule
 & \multicolumn{4}{c}{\textbf{\textbf{Metric}}}\tabularnewline
\cline{2-5}
\textbf{Model}  & \begin{turn}{90}
{\tabular{l}F1 weighted\endtabular}
\end{turn} &
\begin{turn}{90}
{\tabular{l}Prec. weighted\endtabular}
\end{turn} &
\begin{turn}{90}
{\tabular{l}Rec. weighted/Acc\endtabular}
\end{turn} &
\begin{turn}{90}
{\tabular{l}Top 3 Acc\endtabular}
\end{turn}\tabularnewline
\hline
\hline
Base              &        0.494 &           0.478 &     0.527 &     0.805 \\
Base crop         &        0.504 &           0.494 &     0.553 &     \textbf{0.823} \\
NS                &        0.519 &           0.512 &     0.565 &     0.820 \\
NS crop           &        0.534 &           0.525 &     0.586 &     0.822 \\
NS crop no freeze &        0.522 &           0.521 &     0.585 &     0.812 \\
NS no freeze      &        0.517 &           0.504 &     0.555 &     0.816 \\
PT                &        \textbf{0.543} &           \textbf{0.550} &    \textbf{ 0.590} &     0.820 \\
PT crop           &        0.532 &           0.545 &     0.589 &     0.831 \\
PT crop no freeze &        0.512 &           0.511 &     0.551 &     \textbf{0.823} \\
PT no freeze      &        0.499 &           0.485 &     0.526 &     \textbf{0.823} \\
UPT               &        0.484 &           0.484 &     0.563 &     0.794 \\
\midrule
Random forest &        0.539 &           0.533 &     0.579 &     0.813 \\
\bottomrule
\end{tabular}
\label{tab:nat_avg_table}
\end{table}

Best performance, measured by the weighted F1-score, was achieved using a plain transfer learning model, pretrained with the CORINE land cover labels. No cropping or semi-supervised learning was used for this model, however further studies presented later show that cropping and semi-supervised learning can help classification in the general case. The overall performance of the baseline random forest model is better than any of the ResNet models alone, but ensembling these models with the random forest model boosts performance significantly.

\begin{figure}[htb]
     \centering
     \begin{subfigure}{.48\linewidth}
        \centering
        \includegraphics[width=1\textwidth]{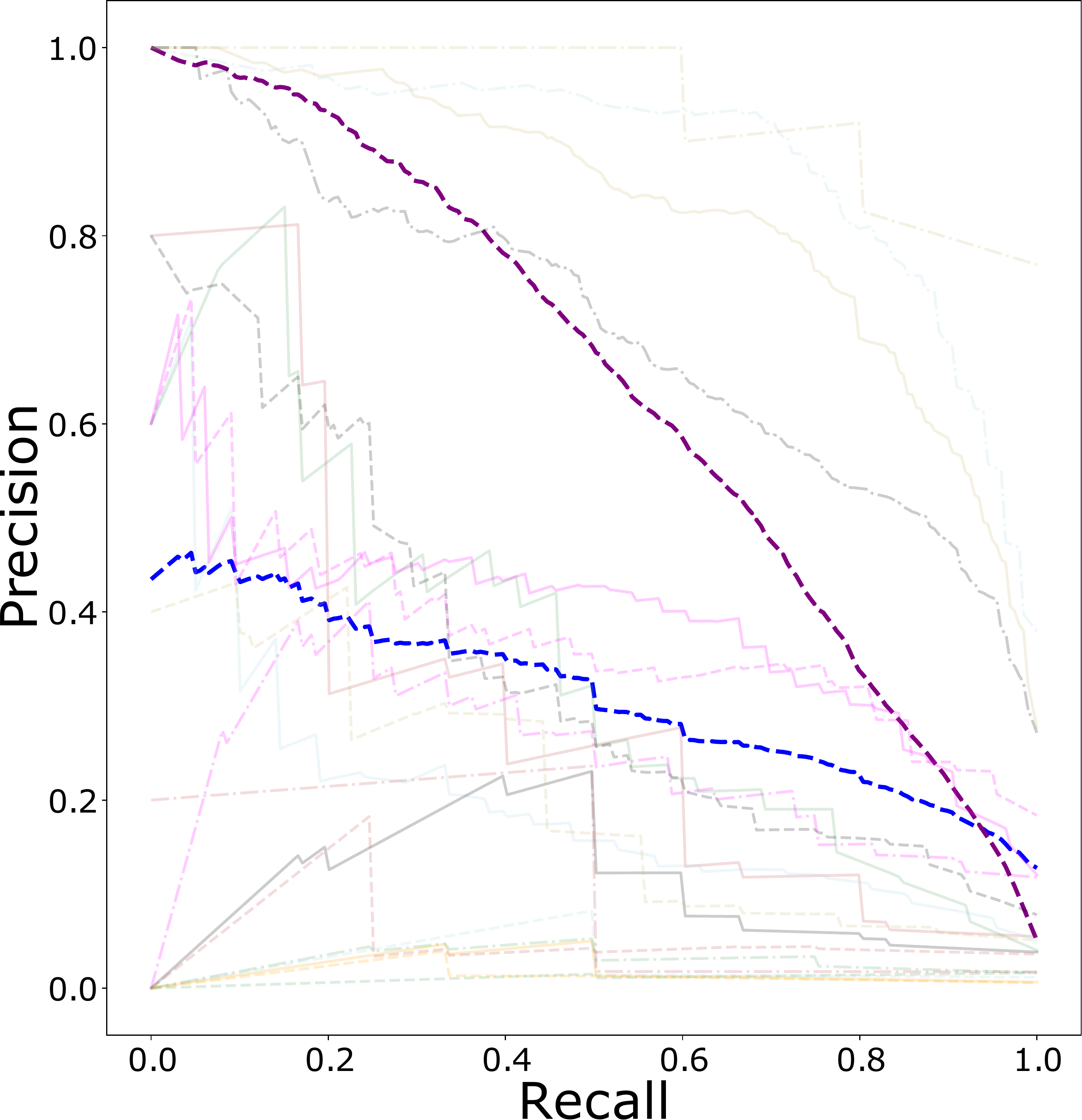}
        \caption{Random forest}
        \label{fig:classwise_prec_rec_rf}
     \end{subfigure}\hfill
     \begin{subfigure}{.48\linewidth}
        \centering
        \includegraphics[width=1\textwidth]{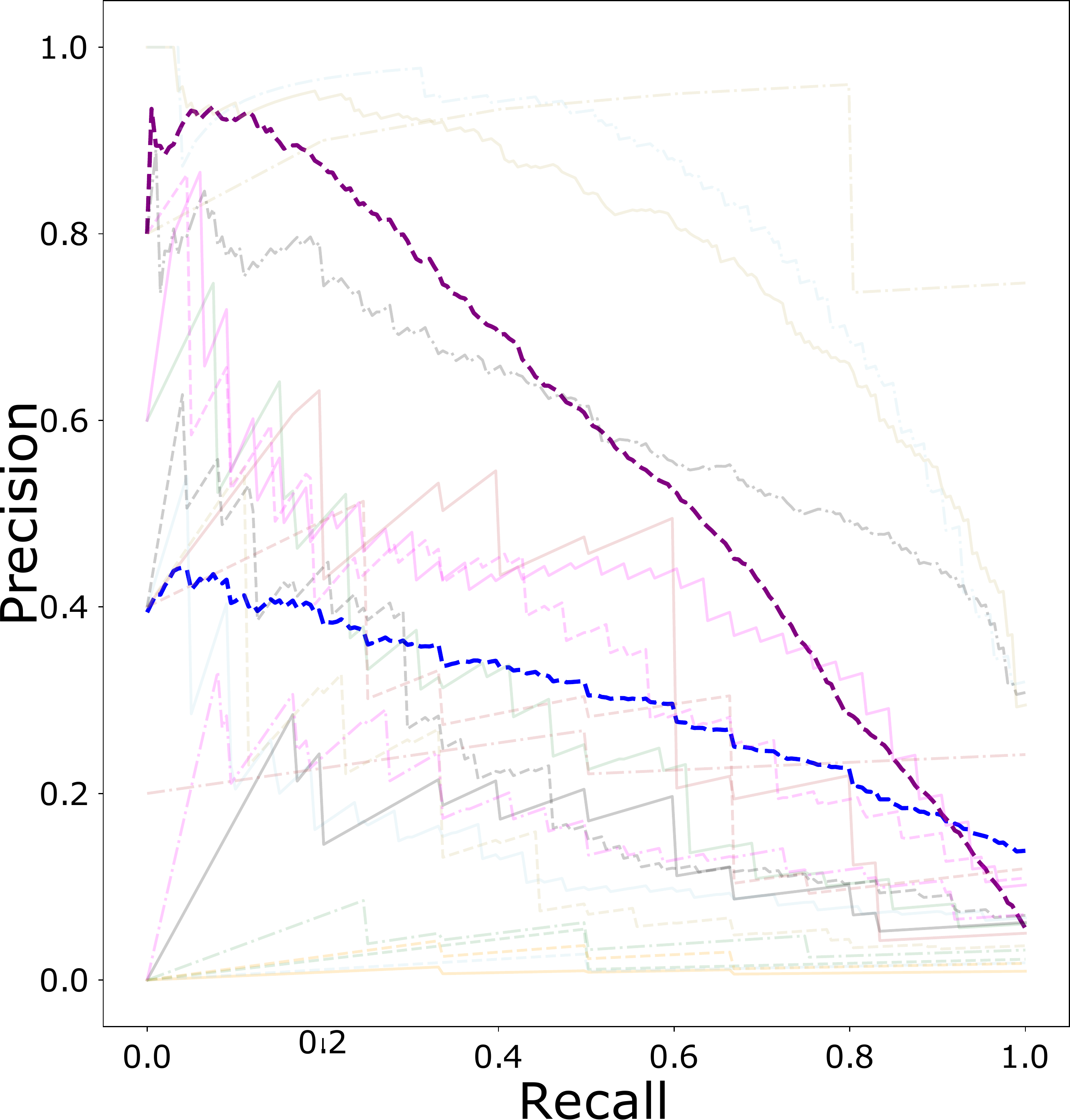}
        \caption{ResNet}
        \label{fig:classwise_prec_rec_resnet}
     \end{subfigure}\\
     \begin{subfigure}{.48\linewidth}
        \centering
        \includegraphics[width=1\textwidth]{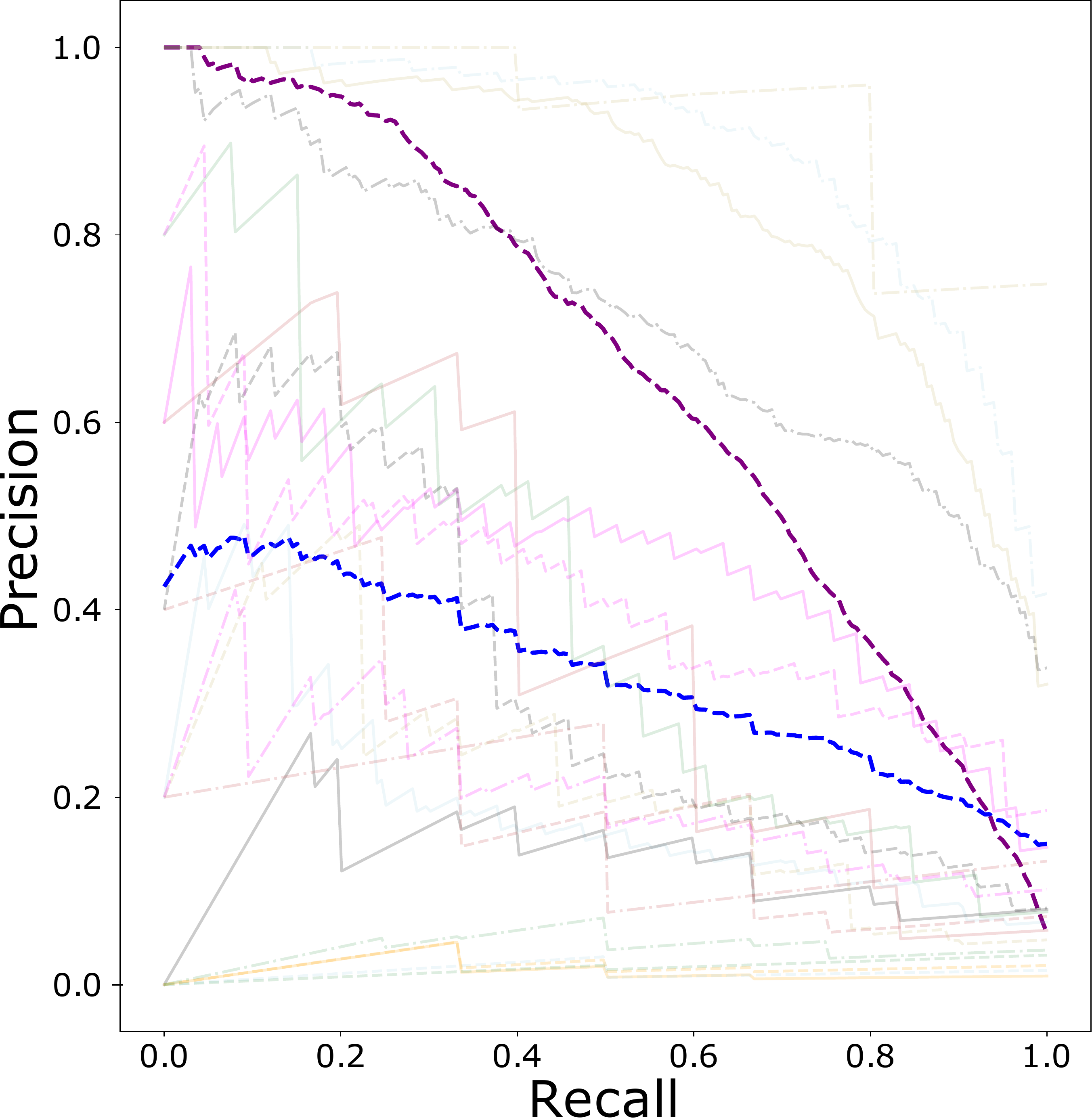}
        \caption{Ensemble model}
        \label{fig:classwise_prec_rec_avg}
     \end{subfigure}
     \begin{subfigure}{.48\linewidth}
        \centering
        \includegraphics[width=1\textwidth]{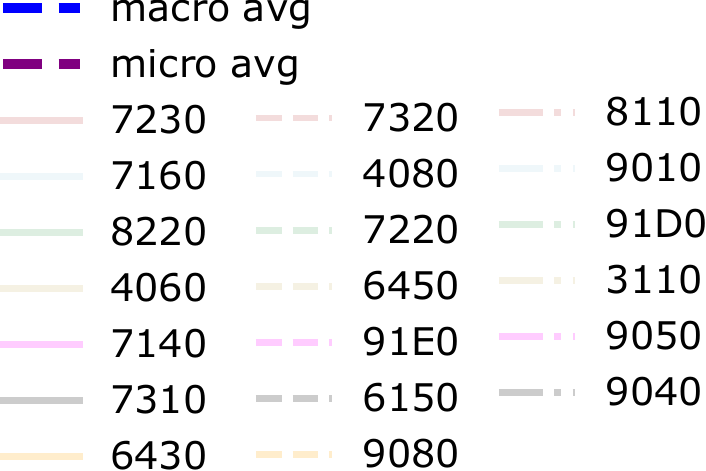}
     \end{subfigure}
     
    \caption{Class-wise precision-recall curves for Natura2000 classifications using the CORINE-pretrained model and test-time augmentation}
    \label{fig:classwise_prec_rec}
\end{figure}

The performance across different classes can be seen in the precision-recall curves in Figure \ref{fig:classwise_prec_rec}. The performance difference between classes is substantial, with more populous classes being more reliably classified. This can be seen also in the difference between the macro- and micro-averages of the classifiers over classes. Due to the larger classes containing most examples, and them being classified mostly correctly, the micro-average performance is higher. Because the dataset contains several classes with only few examples and the classifier performing poorly on these, the macro-averaged classification performance is considerably lower.

\begin{figure}[tb]
     \centering
     \begin{subfigure}{.5\linewidth}
        \centering
        \includegraphics[width=1\textwidth]{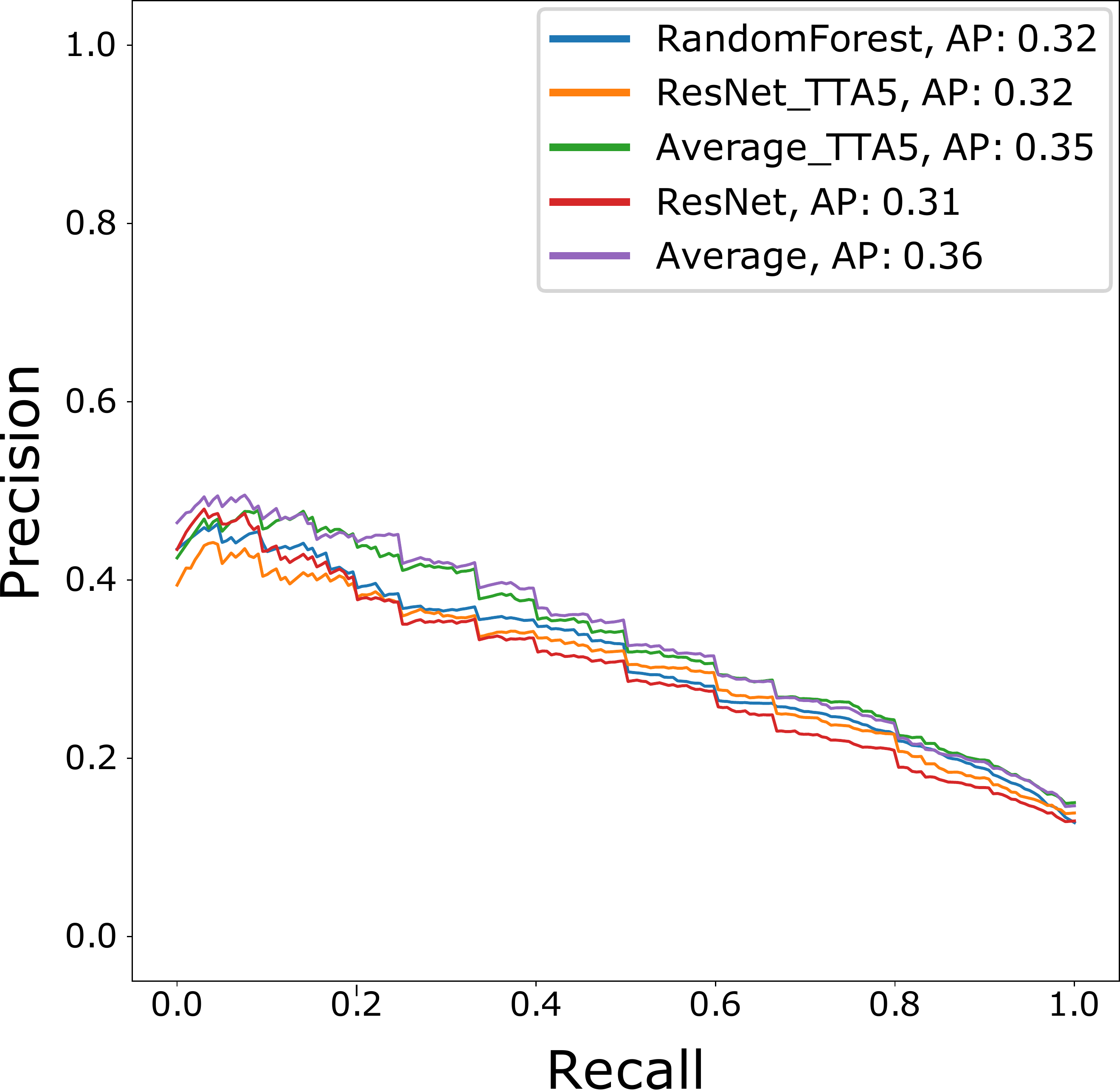}
        \caption{Macro average precision-recall curves}
        \label{fig:curve_precrec_macro}
     \end{subfigure}
     \begin{subfigure}{.5\linewidth}
        \centering
        \includegraphics[width=1\textwidth]{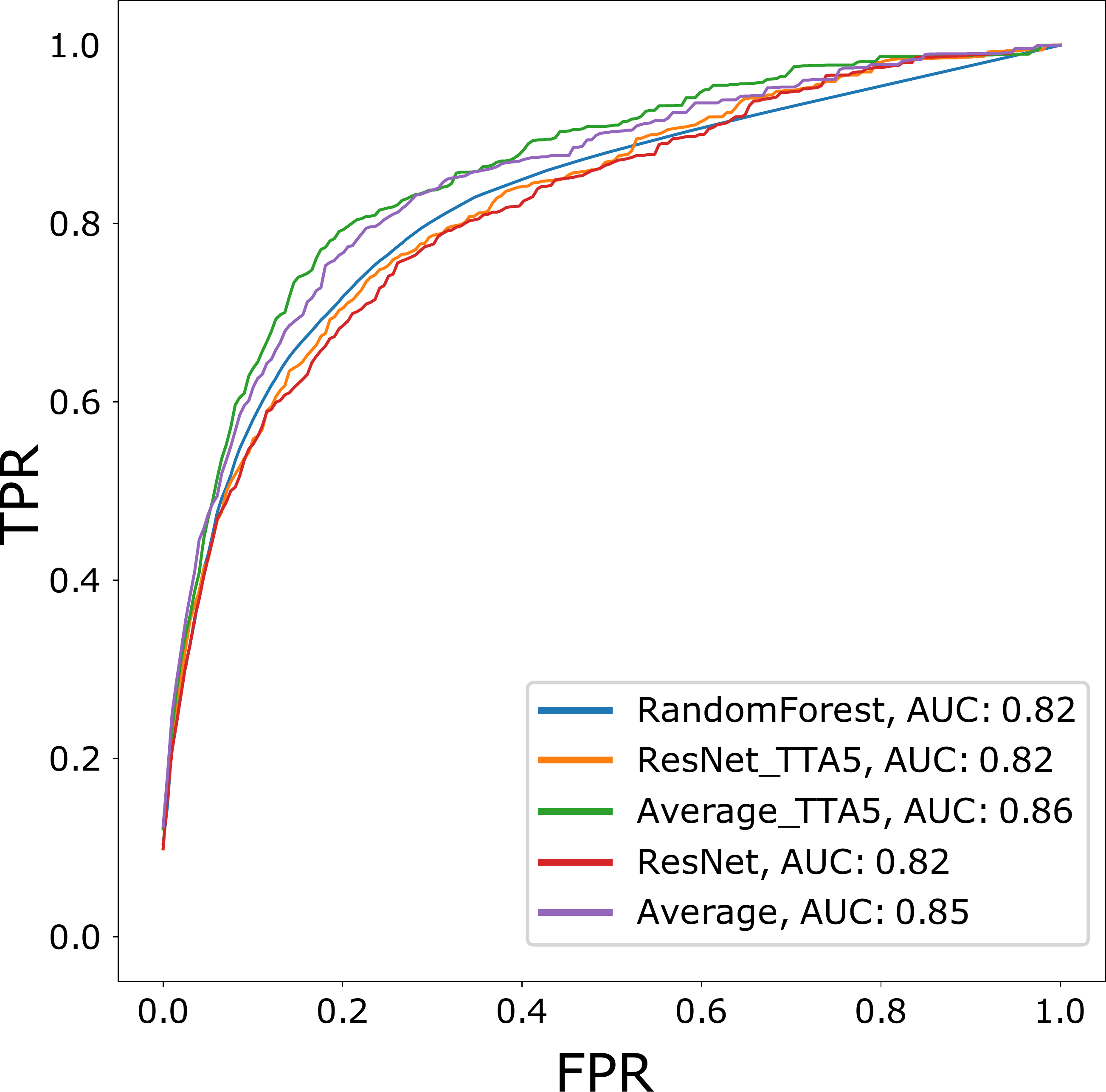}
        \caption{Macro average ROC curves}
        \label{fig:curve_roc_macro}
     \end{subfigure}\hfill
     
    \caption{Natura2000 classes precision-recall and ROC curve comparisons between CORINE-pretrained ResNet, random forest, and ensemble models with test-time augmetation applied five times (TTA5)}
    \label{fig:comparison_roc}
\end{figure}

Figure \ref{fig:comparison_roc} illustrates the differences between different models and the effect of test-time augmentation. As figures \ref{fig:curve_precrec_macro} and \ref{fig:curve_roc_macro} show, the ensemble model performance gain is higher for the macro-averaged models, indicating better performance in smaller classes. The ensemble model gain is smaller in the micro-averaged models, due to better random forest performance in the larger classes.

\begin{figure}[htb]
     \centering
     \begin{subfigure}{0.6\linewidth}
        \centering
        \includegraphics[width=1\textwidth]{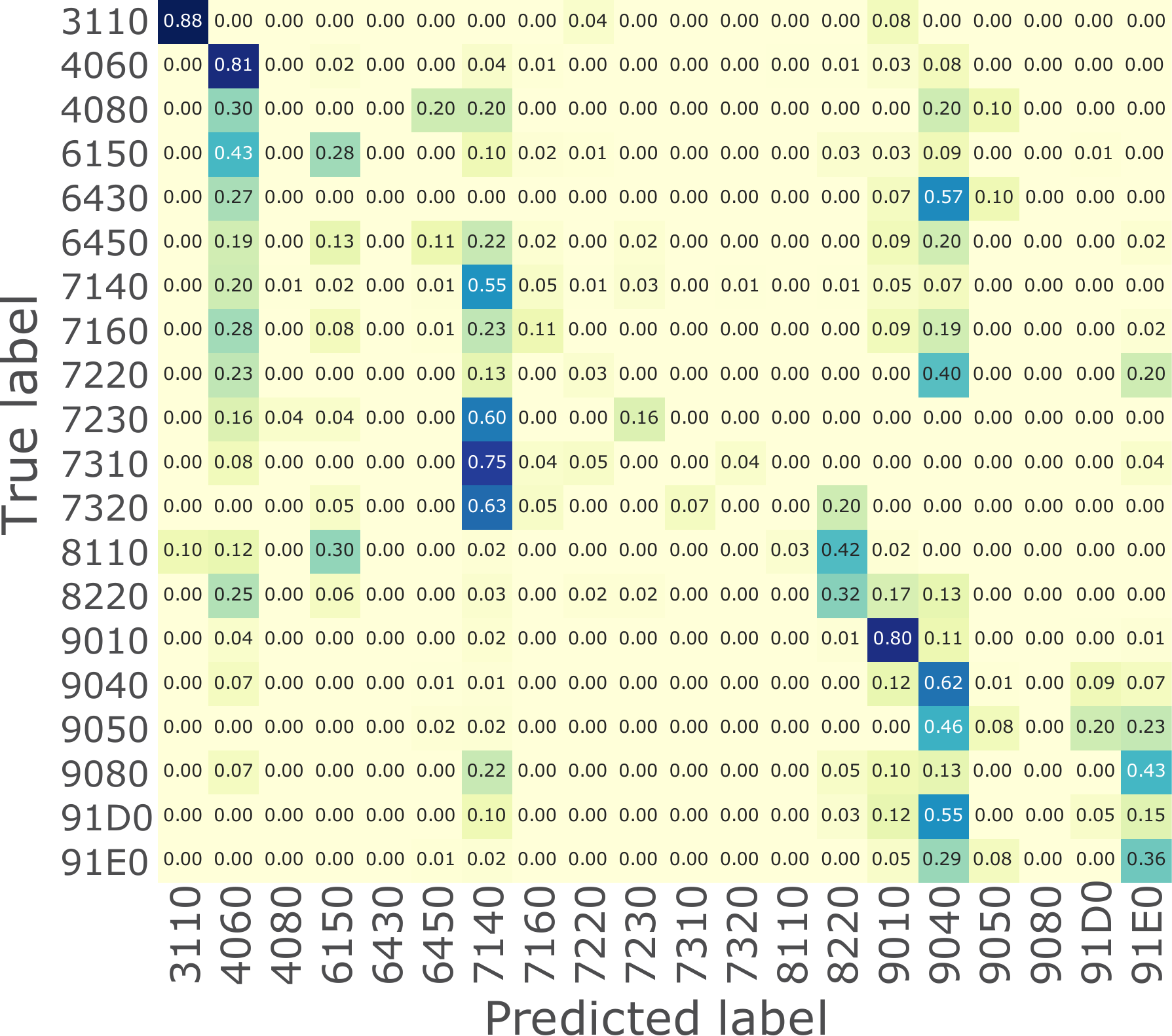}
        \caption{Random forest}
        \label{fig:confmat_rf}
     \end{subfigure}\\
     \begin{subfigure}{0.6\linewidth}
        \centering
        \includegraphics[width=1\textwidth]{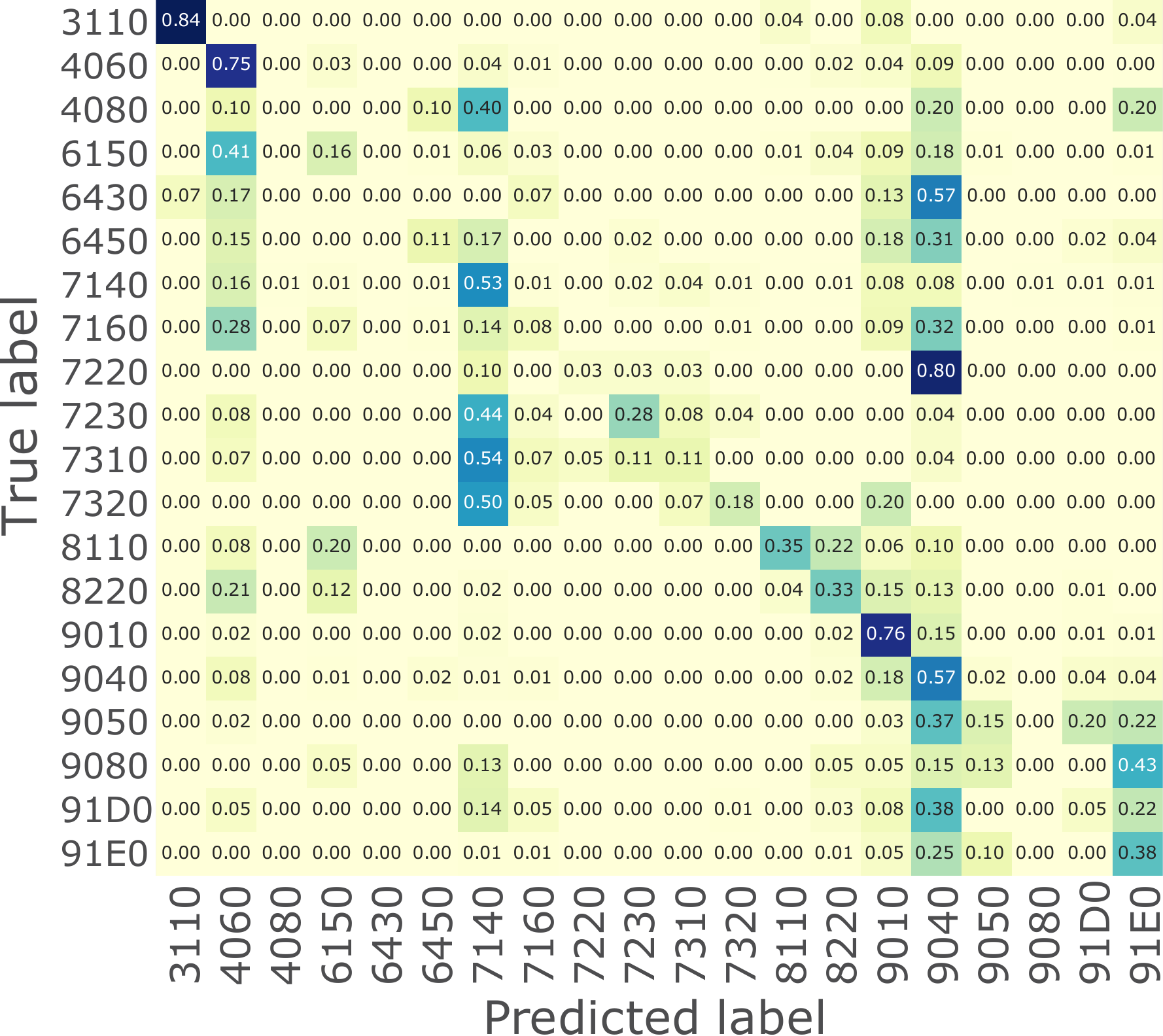}
        \caption{ResNet}
        \label{fig:confmat_resnet}
     \end{subfigure}
     
    \caption{Normalized confusion matrices for Natura2000 classes using CORINE-pretrained models with test-time augmentation applied}
    \label{fig:confmat_nat}
\end{figure}

The confusion matrices in Figure \ref{fig:confmat_nat} are very similar to each other, with concentrating classifications to a few classes. Most different wetland types are classified to the class \textit{"7140 - Transition mires and quaking bogs"}, and smaller classes are often falsely classified to the largest class \textit{"9040 - Nordic subalpine/subarctic forests"}.

\clearpage
\subsection{Sensitivity studies}

Different training methods can have a large effect on the final outcome of the classifier. We tested four different hypotheses:

\begin{enumerate}
    \item Pretraining using a large dataset with a coarse class taxonomy improves classification with fine-grained labels.
    \item Semi-supervised learning with the larger dataset (without labels) improves performance from the pre-trained classifier
    \item Training only the classification head with the fine-grained labels leads to a more generalized model
    \item Augmentation by random center cropping improves performance in this specific scenario.
\end{enumerate}

\begin{figure}[!htb]
     \centering
     \begin{subfigure}{.8\linewidth}
        \centering
        \includegraphics[width=1\textwidth]{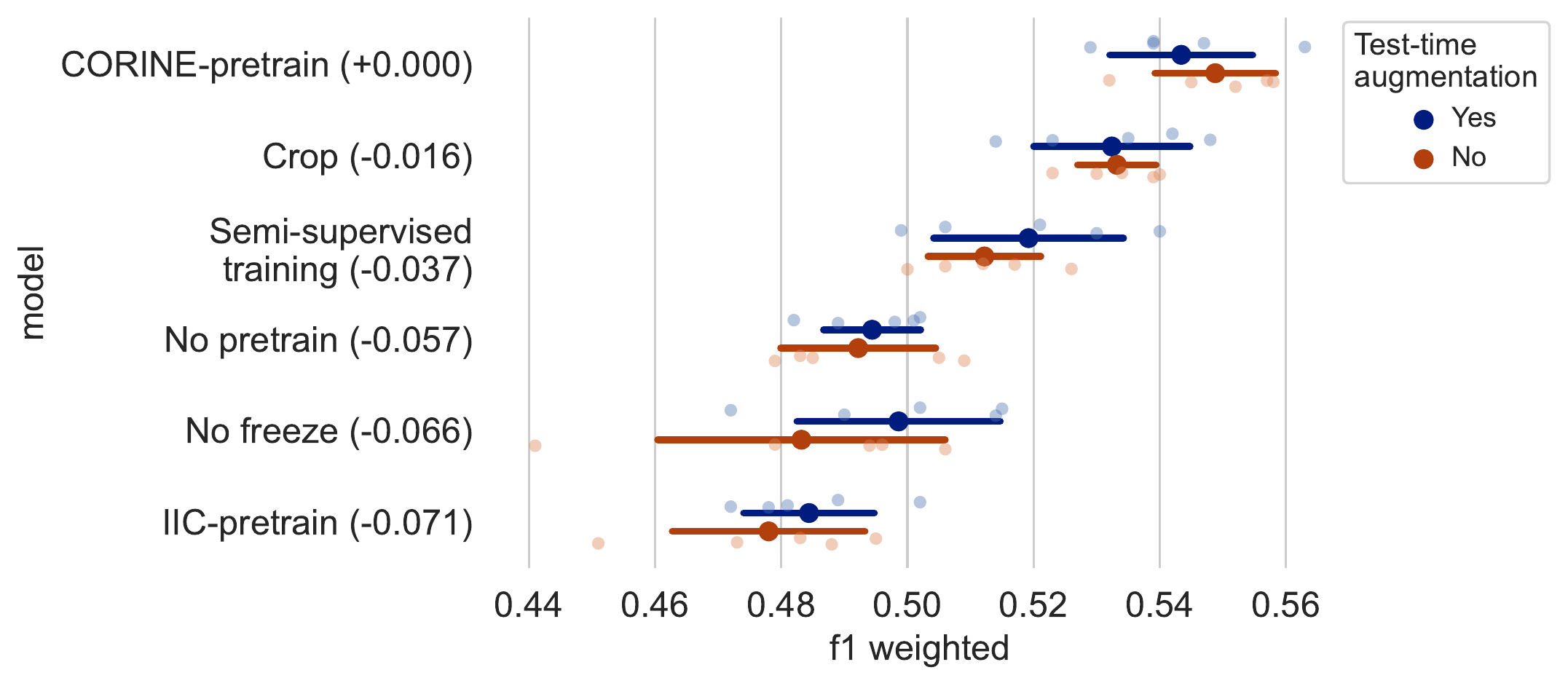}
        \caption{F1 weighted}
        \label{fig:ablation_f1w_nat}
     \end{subfigure}\\
     \begin{subfigure}{.8\linewidth}
        \centering
        \includegraphics[width=1\textwidth]{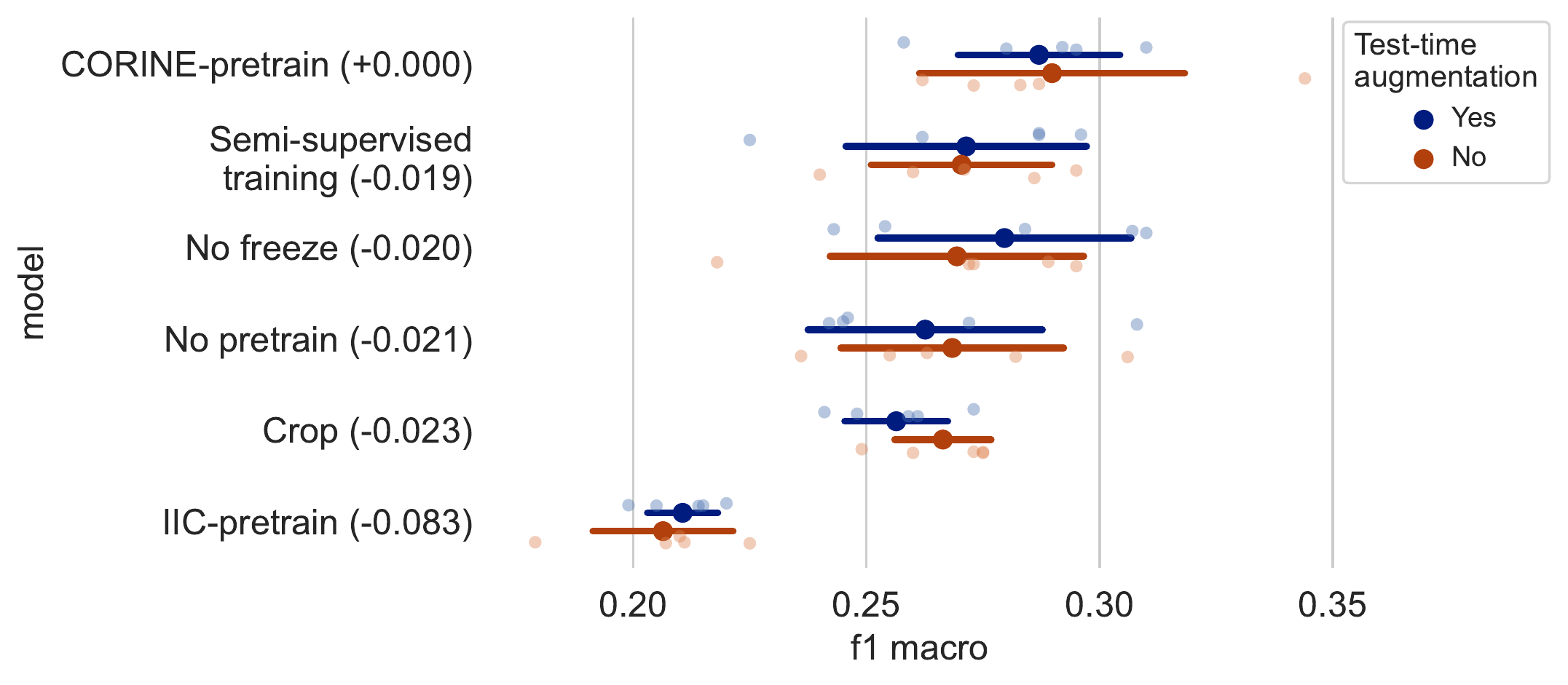}
        \caption{F1 macro}
        \label{fig:ablation_f1m_nat}
     \end{subfigure}
     
    \caption{Sensitivity study for attributes affecting the best Natura2000 CORINE-pretrained model, with test-time augmentation effect plotted separately. Each scatterplot point is a result of a cross-validation fold. Large points are mean of all cross-validation folds, with the standard deviation as a bold line.}
    \label{fig:ablation_nat}
\end{figure}

Figure \ref{fig:ablation_nat} illustrates the effect of different model ablations on the best performing model, which is the plain transfer-learned model with CORINE pretraining, without crop augmentations and with convolutional layers freezed after pretraining. The figure shows the weighted and macro F1 scores of each cross-validation fold, with the mean and single standard deviation range highlighted in bold. The effect of adding a certain attribute to the best performing model is shown under the best model performance. Test-time augmentation can be applied for each model at test time, and is plotted separately for each attribute.

\begin{figure}[!htb]
     \centering
     \begin{subfigure}{.8\linewidth}
        \centering
        \includegraphics[width=1\textwidth]{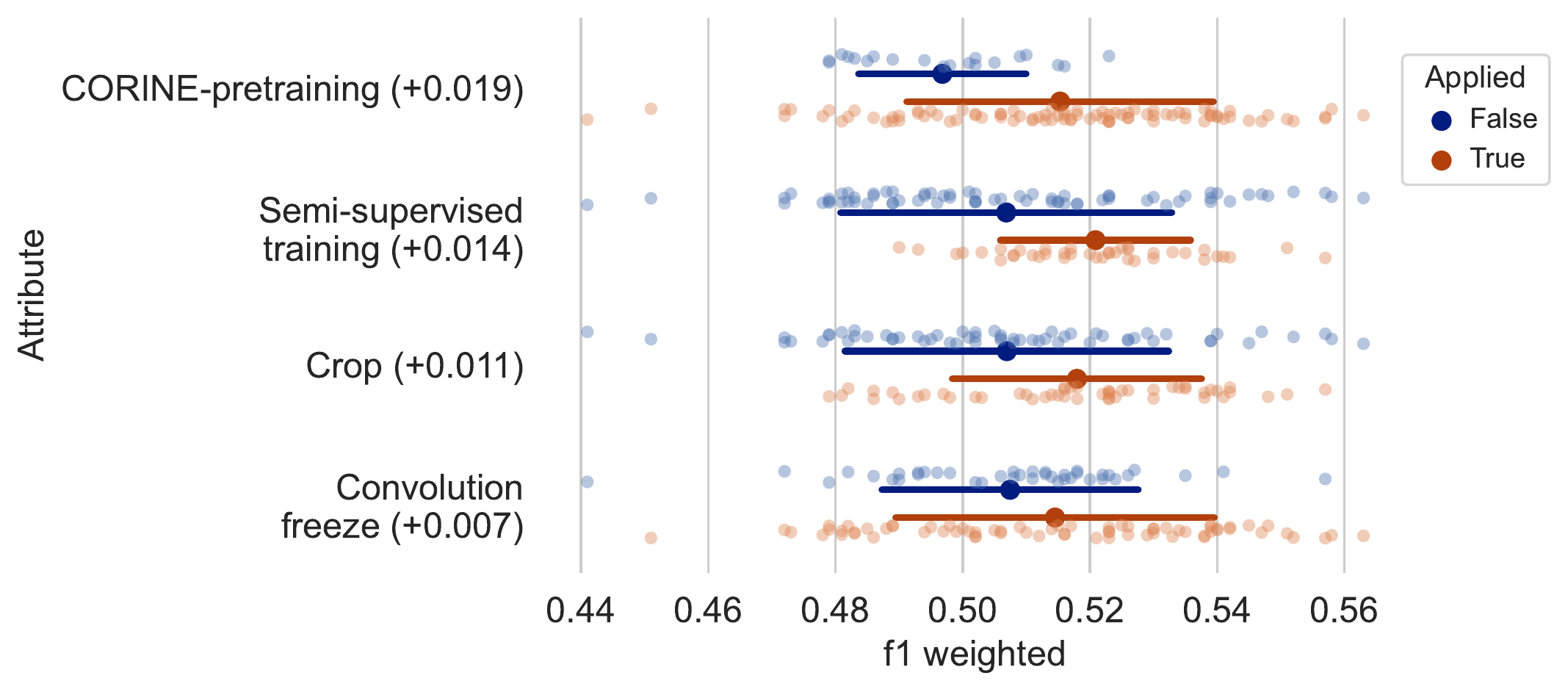}
        \caption{F1 weighted}
        \label{fig:ablation_attribute_f1w_nat}
     \end{subfigure}\\
     \begin{subfigure}{.8\linewidth}
        \centering
        \includegraphics[width=1\textwidth]{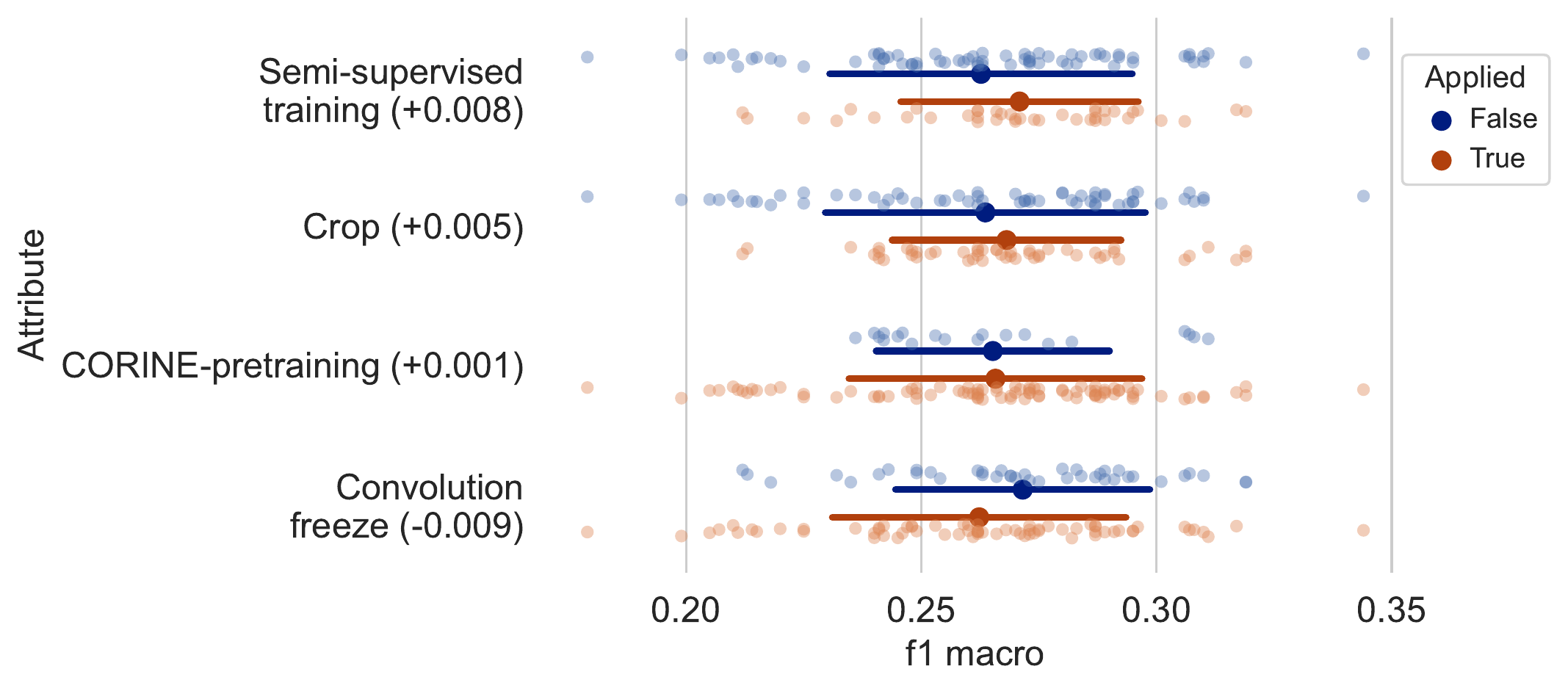}
        \caption{F1 macro}
        \label{fig:ablation_attribute_f1m_nat}
     \end{subfigure}
     
    \caption{Natura2000 training attribute comparison for each attribute separately, with the improvement between attribute being applied or not. Each scatterplot point is a result of a cross-validation fold. Large points are mean of all cross-validation folds, with the standard deviation as bold line.}
    \label{fig:ablation_attribute_nat}
\end{figure}

The effect of pretraining can be also seen in Figure \ref{fig:ablation_attribute_nat}, where the effect of each attribute is plotted separately against all of the other models. Mean and standard deviation over the cross-validation folds of all models where an attribute is used, are plotted as well as the differences between having the attribute and not having it.

If all models with a certain attribute are taken into account, all four methods (pretraining, convolutional layer freezing, semi-supervised learning and cropping) increase the weighted F1 score slightly. For the macro-averaged F1 score the difference is smaller, accompanied with a high overlap in standard deviation. Also, cropping has a negative effect on macro-averaged F1 score.

\clearpage

\subsection{Classification maps}

\begin{figure}[htb]
     \centering
     \begin{subfigure}{.48\linewidth}
        \centering
        \includegraphics[width=1\textwidth]{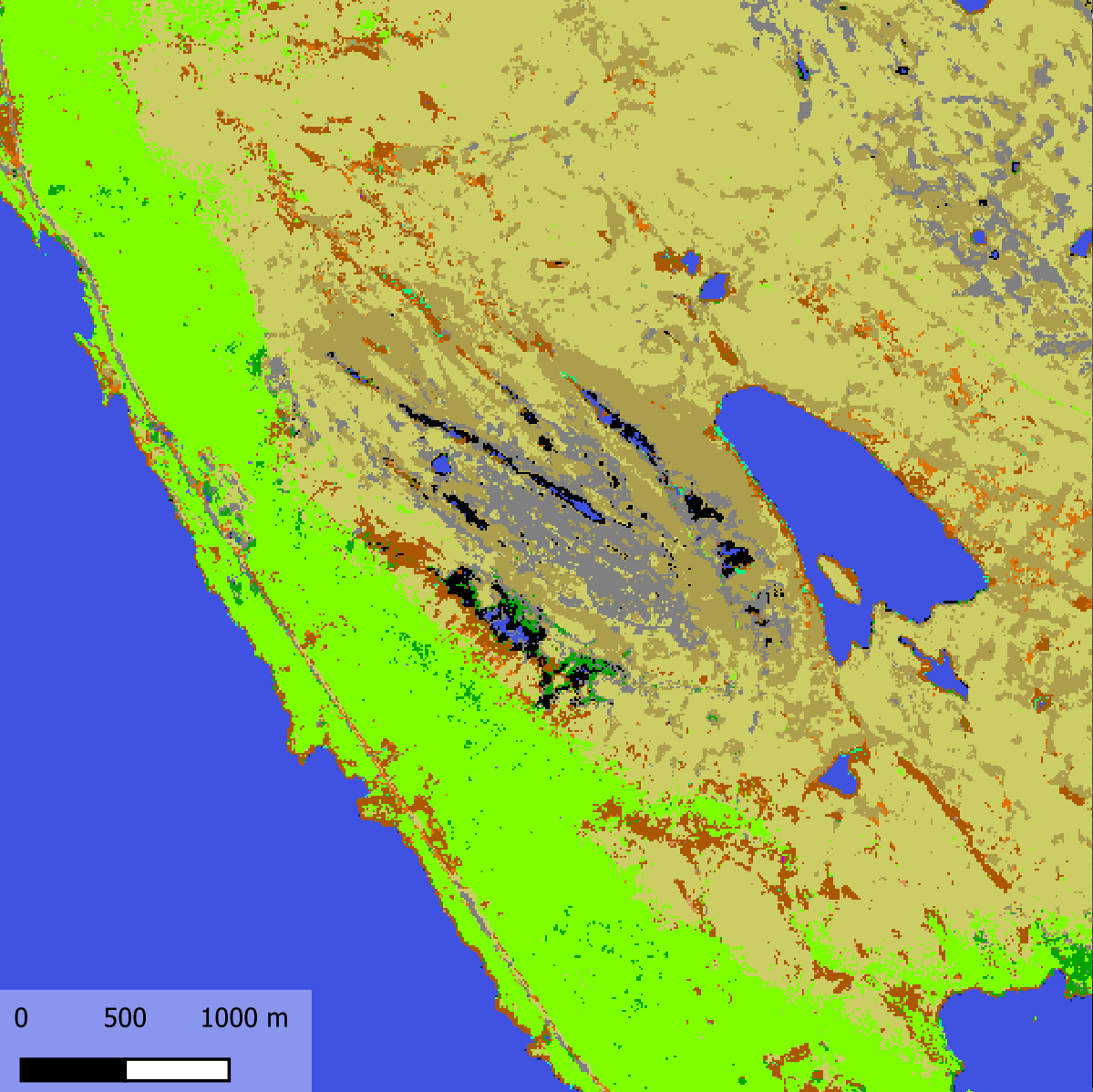}
        \caption{Random forest Natura2000}
        \label{fig:class_saana_nat_rf}
     \end{subfigure}
     \begin{subfigure}{.48\linewidth}
        \centering
        \includegraphics[width=1\textwidth]{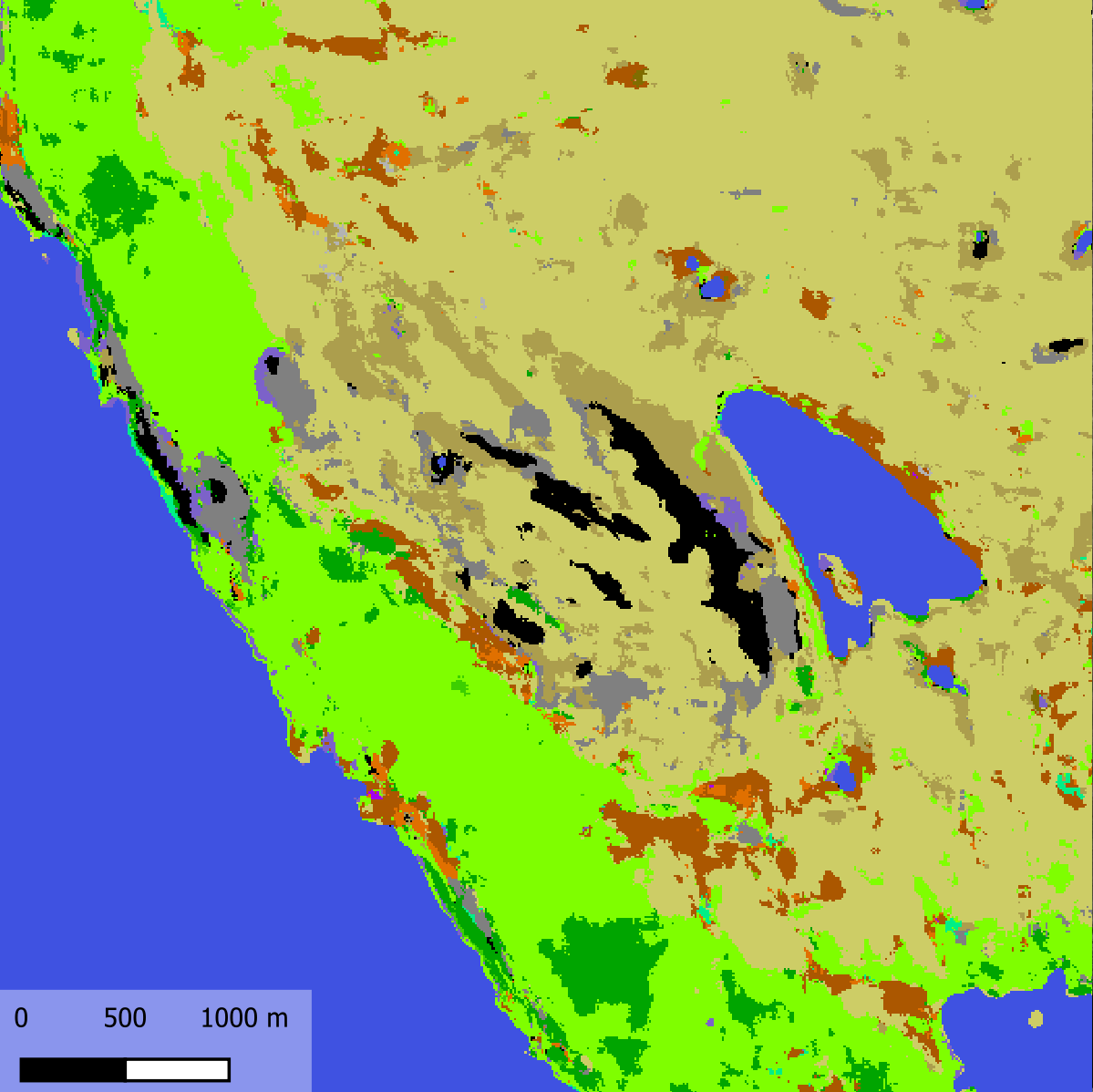}
        \caption{ResNet Natura2000}
        \label{fig:class_saana_nat_resnet}
     \end{subfigure}\\
     \begin{subfigure}{.48\linewidth}
        \centering
        \includegraphics[width=1\textwidth]{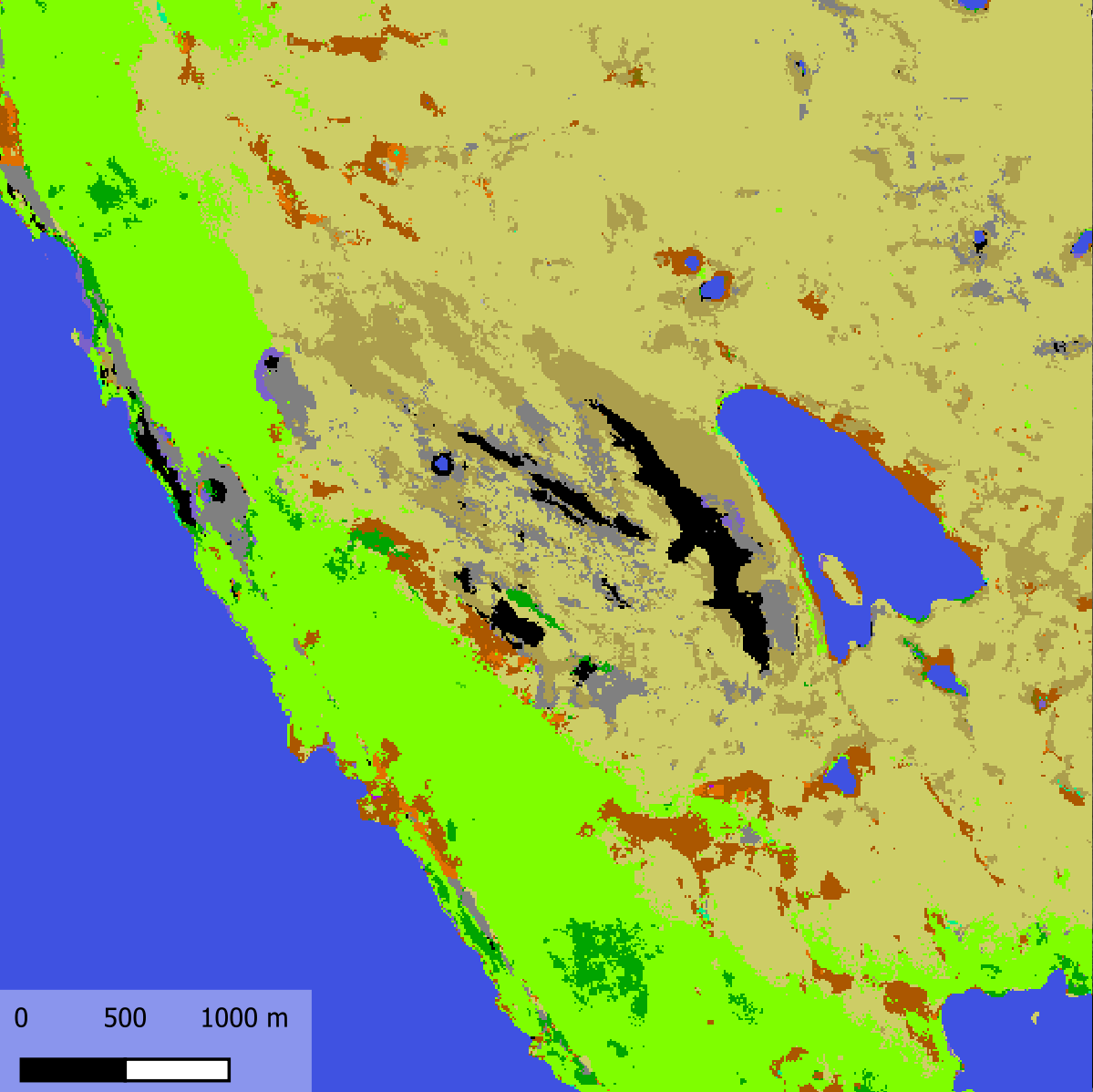}
        \caption{Ensemble model Natura2000}
        \label{fig:class_saana_nat_avg}
     \end{subfigure}
     
    \caption{Classification maps of the Saana fell area for all model types and class taxonomies.}
    \label{fig:class_saana}
\end{figure}

\begin{figure}[htb]
     \centering
     \begin{subfigure}{.3\linewidth}
        \centering
        \includegraphics[width=1\textwidth]{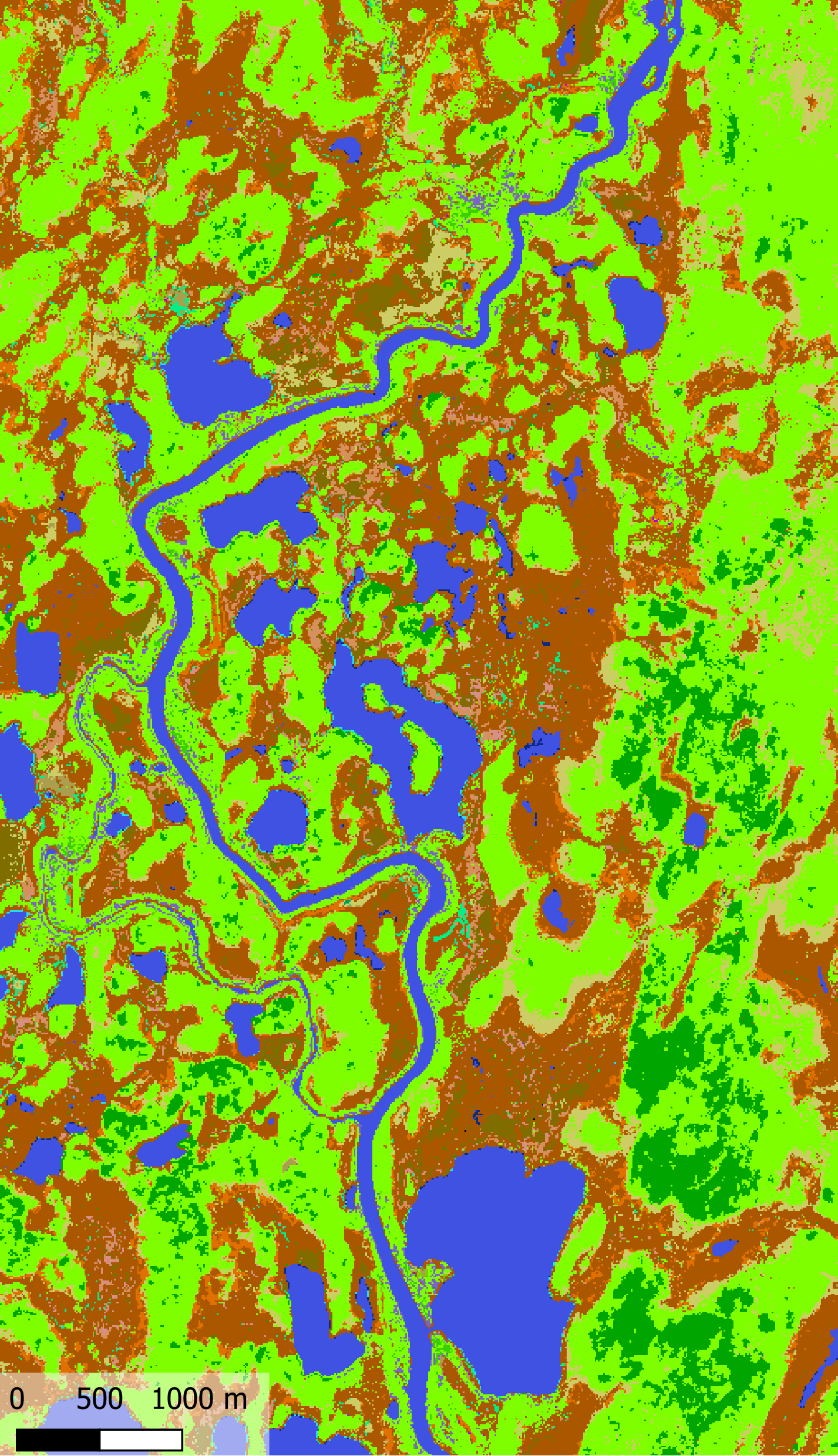}
        \caption{Random forest}
        \label{fig:class_river_nat_rf}
     \end{subfigure}\hfill
     \begin{subfigure}{.3\linewidth}
        \centering
        \includegraphics[width=1\textwidth]{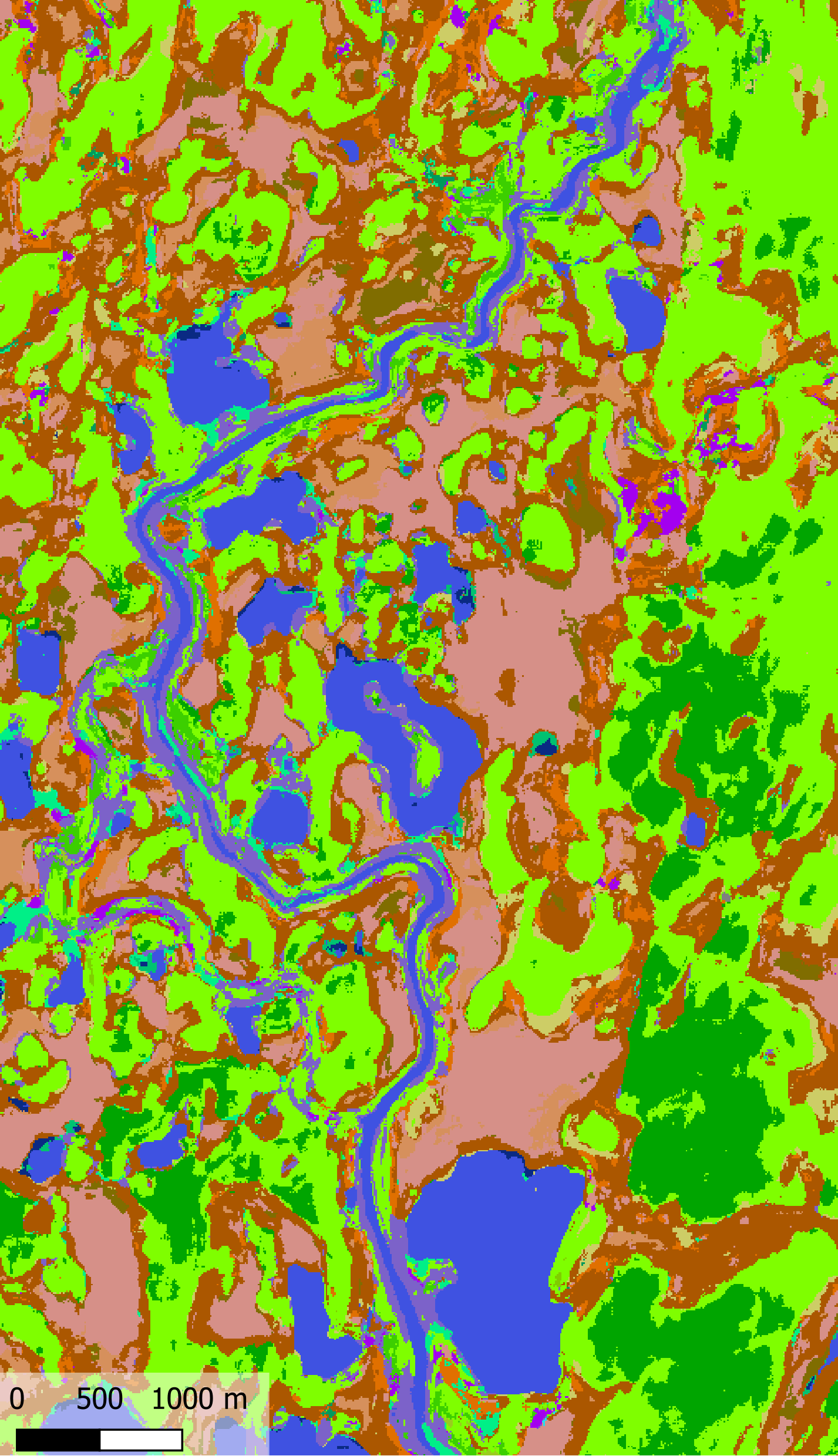}
        \caption{ResNet}
        \label{fig:class_river_nat_resnet}
     \end{subfigure}\hfill
     \begin{subfigure}{.3\linewidth}
        \centering
        \includegraphics[width=1\textwidth]{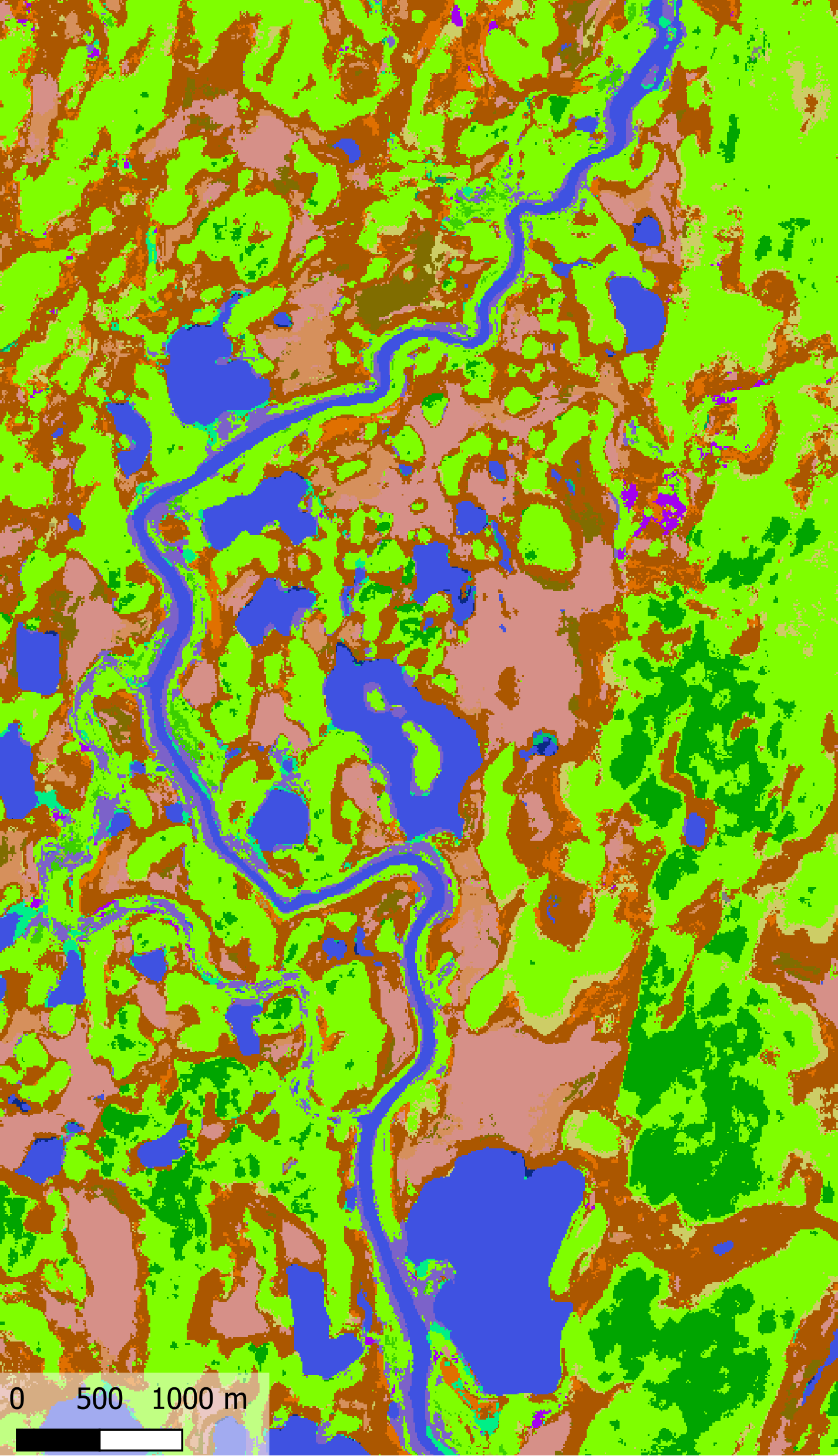}
        \caption{Ensemble model}
        \label{fig:class_river_nat_avg}
     \end{subfigure}
     
    \caption{Classification maps of an area near Lätäseno river for all model types.}
    \label{fig:class_river}
\end{figure}

\begin{figure}[htb]
     \centering
     \begin{subfigure}{.3\linewidth}
        \centering
        \includegraphics[width=1\textwidth]{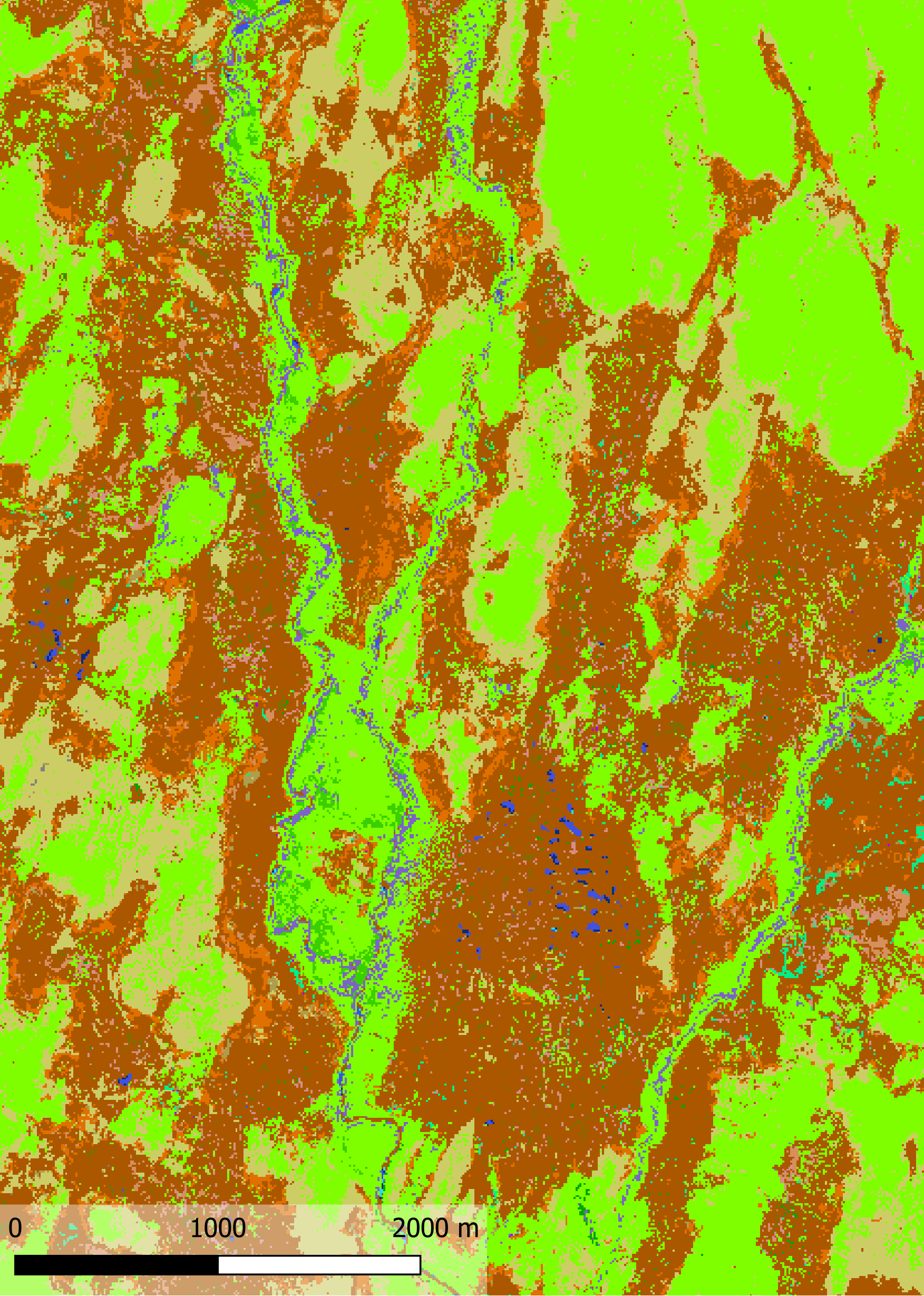}
        \caption{Random forest}
        \label{fig:class_suot_nat_rf}
     \end{subfigure}\hfill
     \begin{subfigure}{.3\linewidth}
        \centering
        \includegraphics[width=1\textwidth]{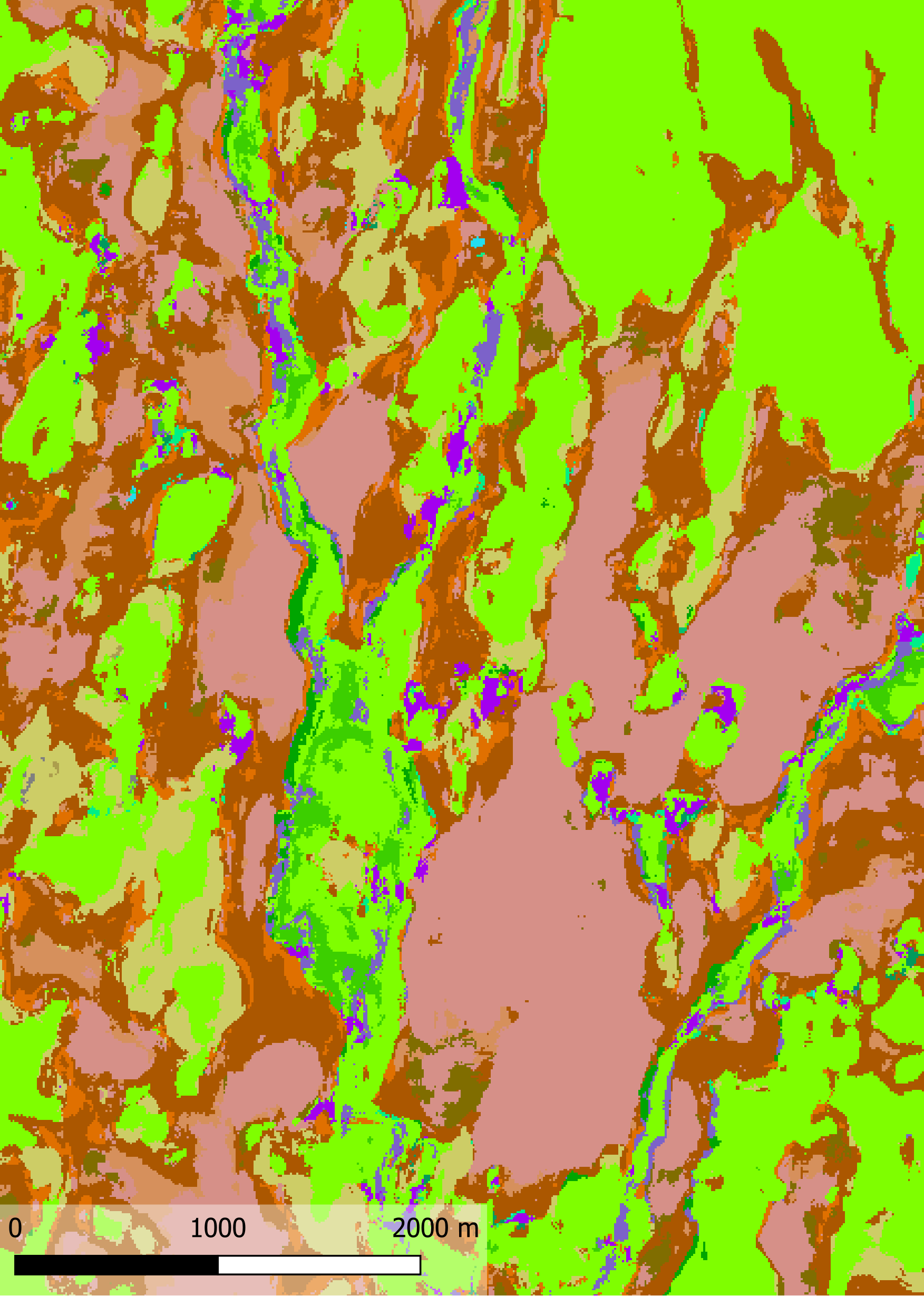}
        \caption{Random forest}
        \label{fig:class_suot_nat_resnet}
     \end{subfigure}\hfill
     \begin{subfigure}{.3\linewidth}
        \centering
        \includegraphics[width=1\textwidth]{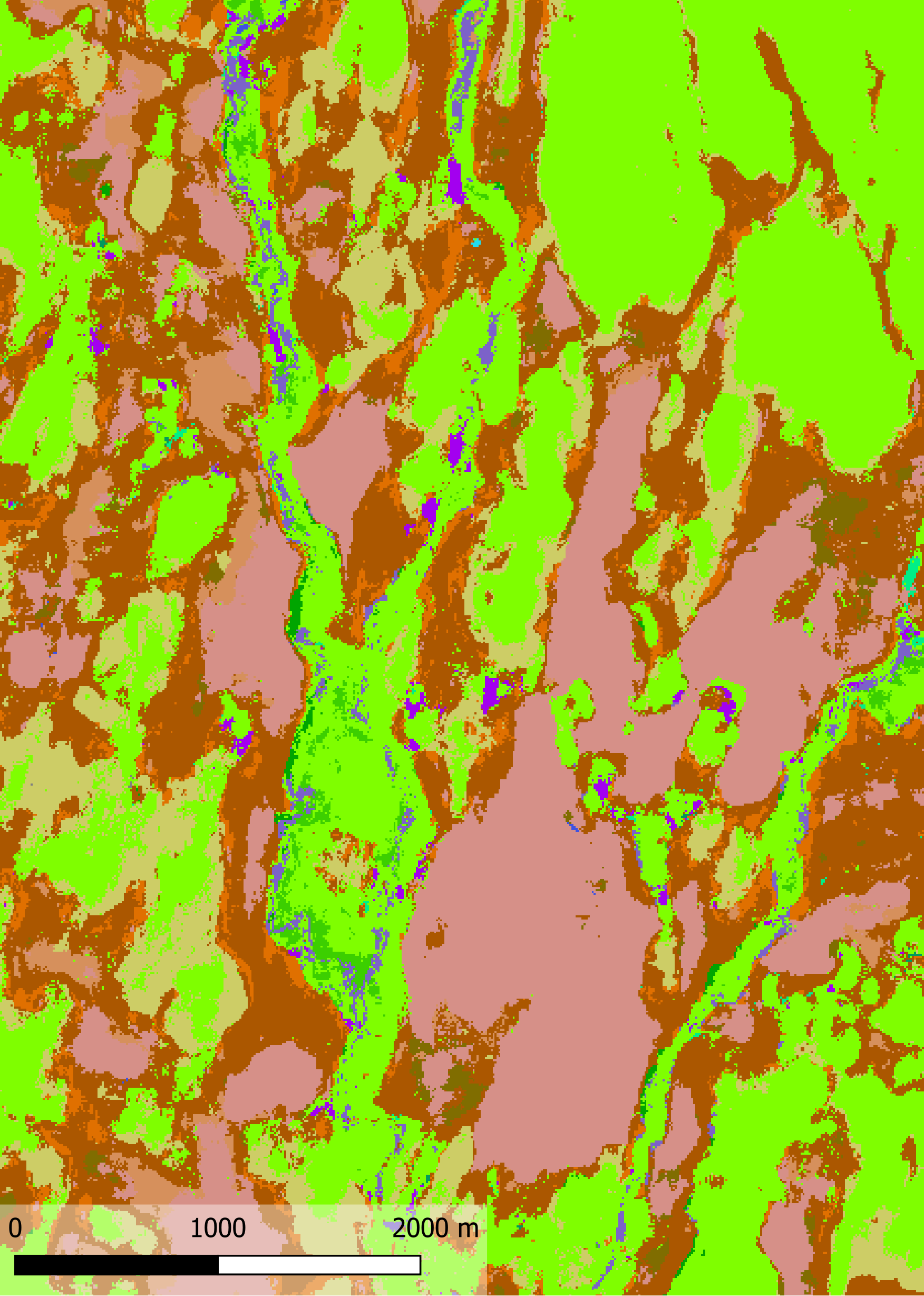}
        \caption{Ensemble model}
        \label{fig:class_suot_nat_avg}
     \end{subfigure}

    \caption{Close-up classification maps of an wetland area near Lätäseno river for all model types.}
    \label{fig:class_suot}
\end{figure}

We produced land cover maps for the entire Northern Lapland study area (extent seen in Figure \ref{fig:lapland}), using the models presented in this paper. Examples of classification maps for different models are displayed in Figures \ref{fig:class_saana}, \ref{fig:class_river}, and \ref{fig:class_suot}. The differences between the models are quite large, the main difference being the more uniform classification map of the ResNet model compared to the fragmentation in the Random Forest's map. Ensembling these models combines these features with some tradeoff.

The models can output not just classification class, but a heuristic probability distribution over all classes in the taxonomy. Using this probability distribution, it is possible to map a single class' classification confidence over the whole study area. This information can be valuable when researchers are trying to find rare nature types from new areas. Even lower confidences of a class might indicate presence in the area. 

\begin{figure}[htb]
     \centering
     \begin{subfigure}{.48\linewidth}
        \centering
        \includegraphics[width=1\textwidth]{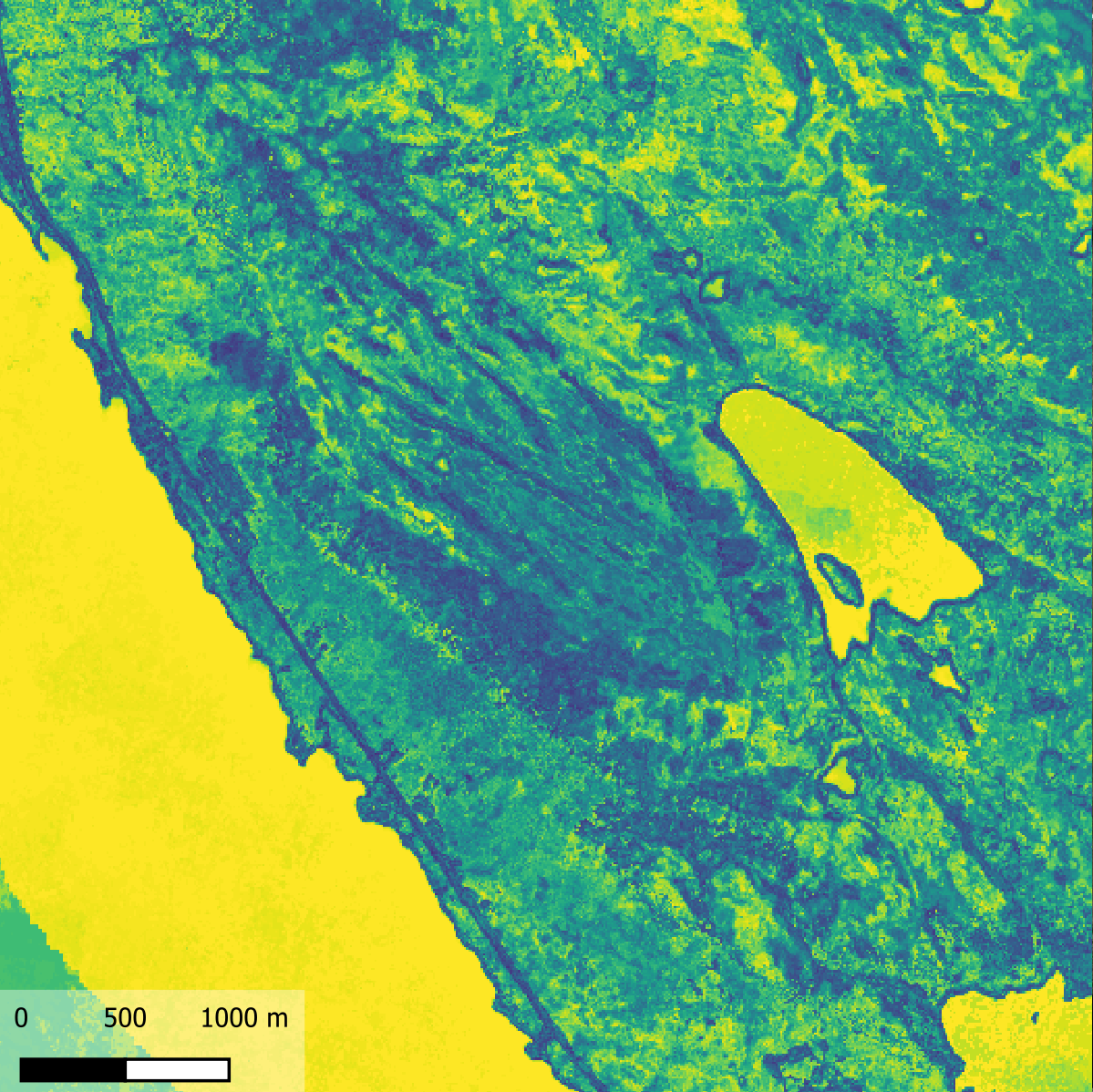}
        \caption{Random forest}
        \label{fig:conf_nat_rf}
     \end{subfigure}\hfill
     \begin{subfigure}{.48\linewidth}
        \centering
        \includegraphics[width=1\textwidth]{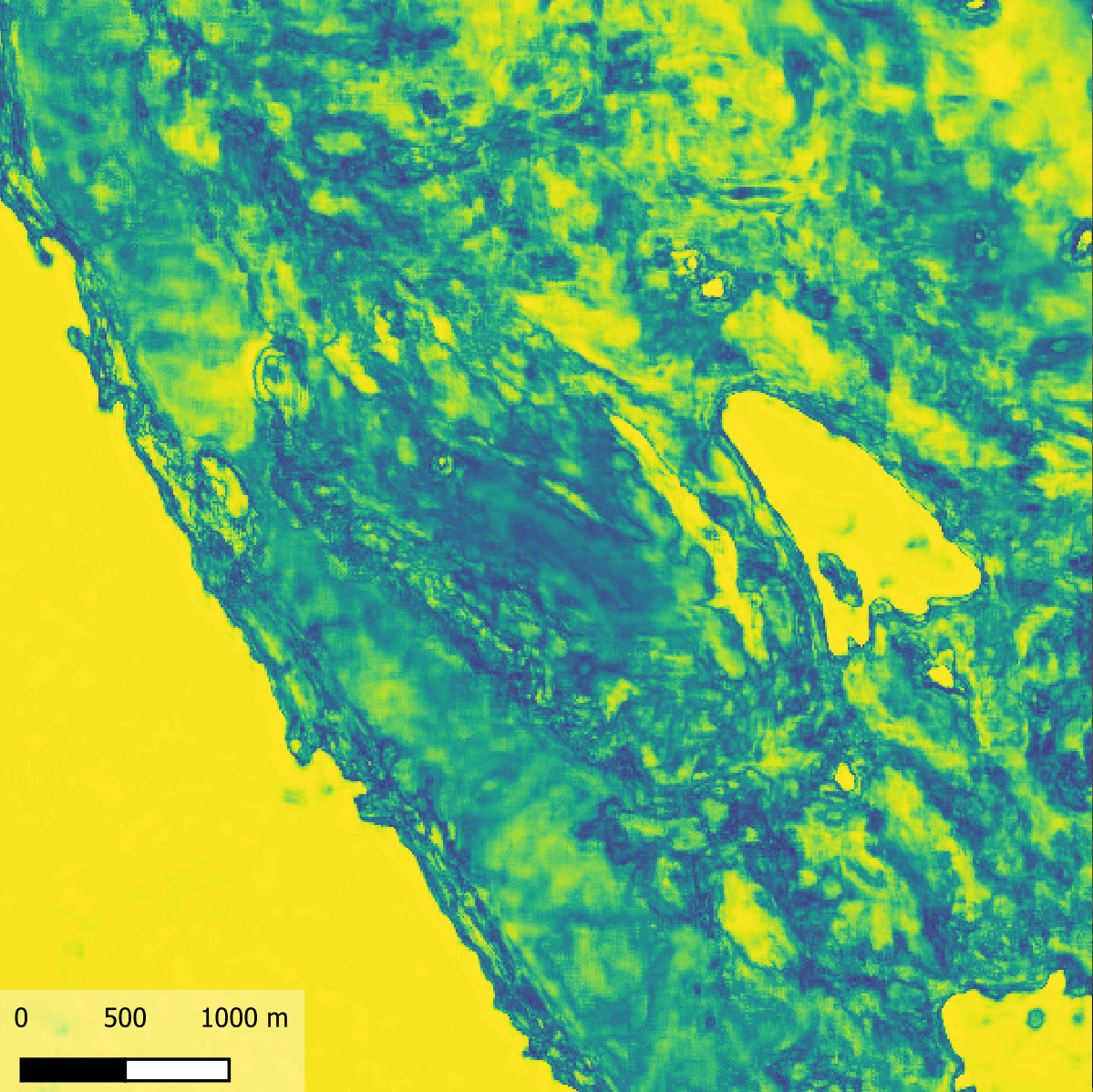}
        \caption{ResNet}
        \label{fig:conf_nat_resnet}
     \end{subfigure}
    \caption{Maximum class heuristic confidence maps of the Saana fell area}
    \label{fig:confidence}
\end{figure}

The class with the maximum confidence is chosen as the final classification for each pixel. In pixels that are harder to classify, the confidence is more distributed among classes, and the maximum class gets a lower confidence score. Mapping the maximum confidence for each pixel, like in Figure \ref{fig:confidence}, it is possible to compare areas where the model is more confident with areas where confidence is low.

\begin{figure}[htb]
     \centering
     \begin{subfigure}{.3\linewidth}
        \centering
        \includegraphics[width=1\textwidth]{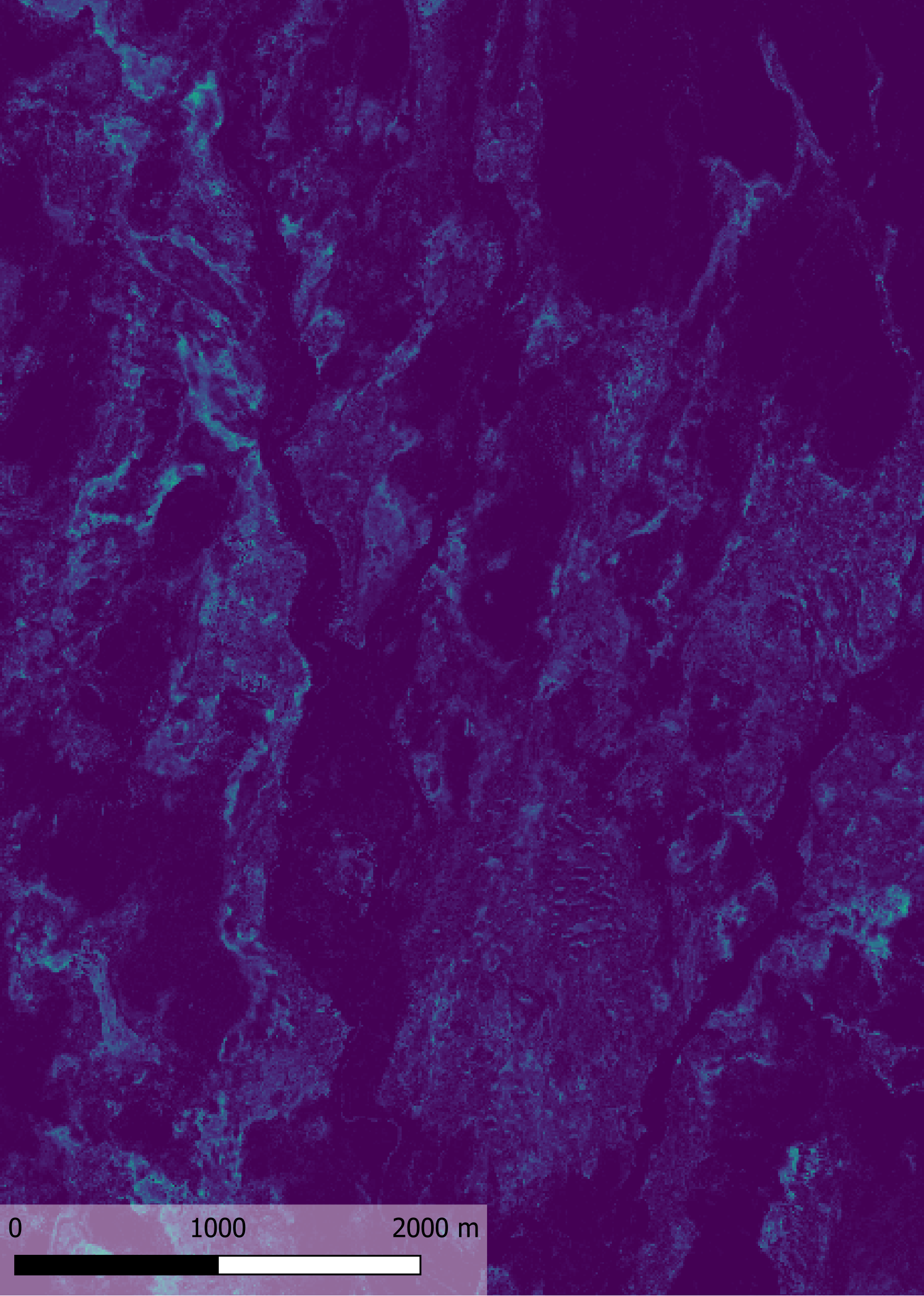}
        \caption{Random forest}
        \label{fig:score_suot_inv_rf}
     \end{subfigure}
     \begin{subfigure}{.3\linewidth}
        \centering
        \includegraphics[width=1\textwidth]{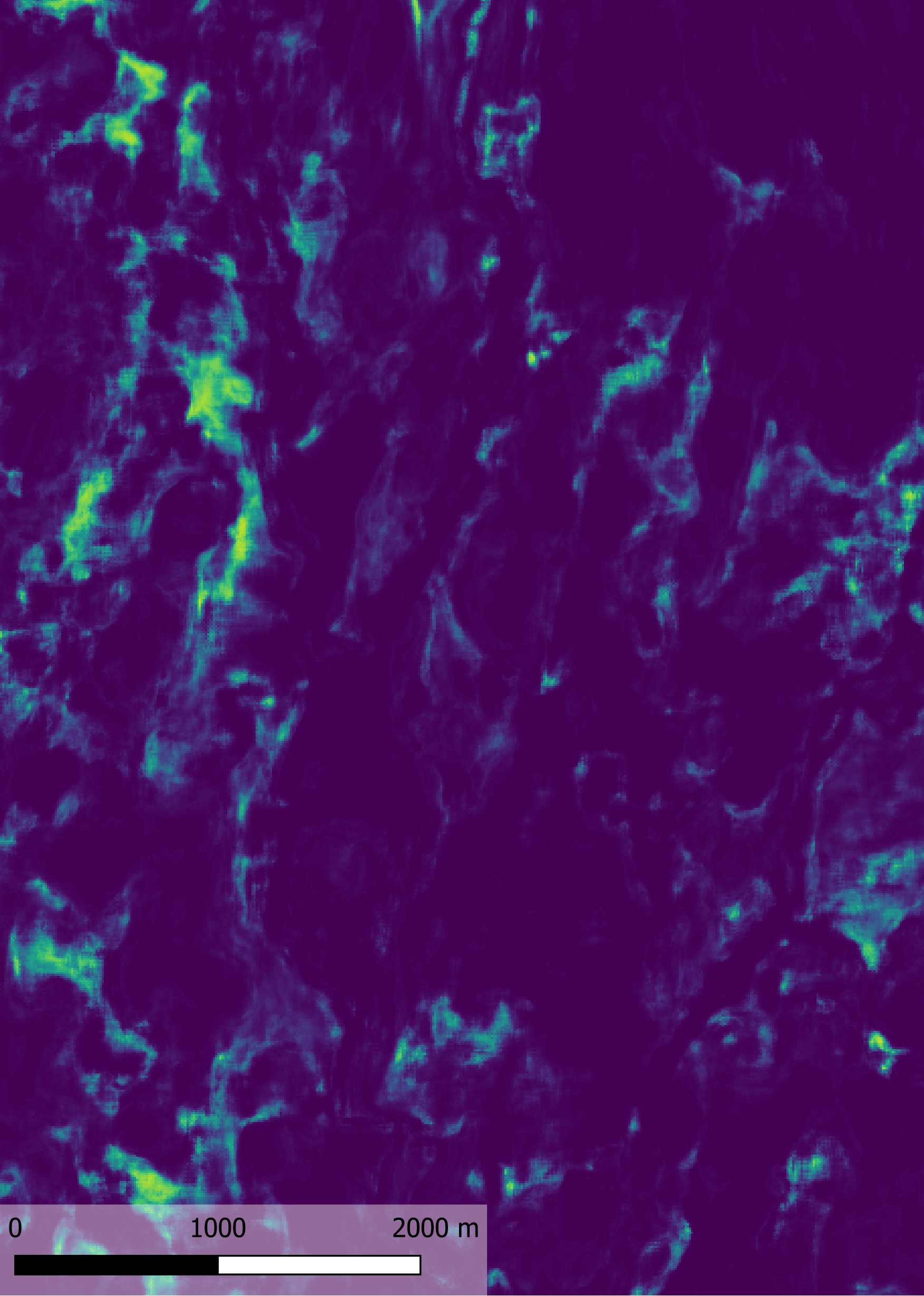}
        \caption{ResNet}
        \label{fig:score_suot_inv_resnet}
     \end{subfigure}\\
     \begin{subfigure}{.3\linewidth}
        \centering
        \includegraphics[width=1\textwidth]{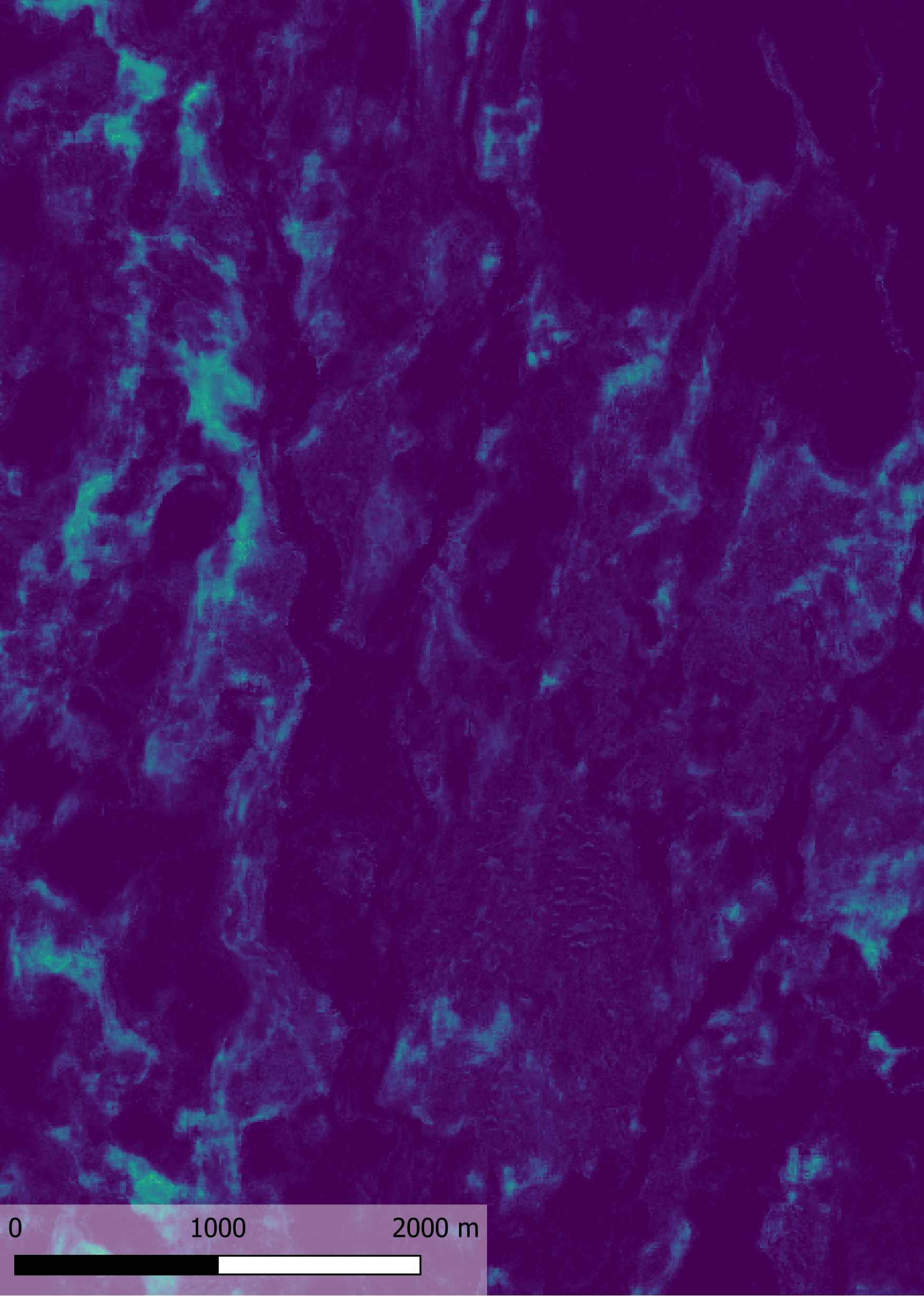}
        \caption{Ensemble model}
        \label{fig:score_suot_inv_avg}
     \end{subfigure}
     \begin{subfigure}{.3\linewidth}
        \centering
        \includegraphics[width=1\textwidth]{class_suot_nat_avg.png}
        \caption{Reference classification}
        \label{fig:score_suot_inv_ref}
     \end{subfigure}
     
    \caption{Classification heatmaps for the Natura2000 class "7310 Aapa mires" at wetland area east of Lätäseno river, with the ensemble classification map as reference.}
    \label{fig:score_suot}
\end{figure}

\clearpage

\section{Discussion}
\label{sec:discussion}

\subsection{Quantitative evaluation}

Overall best performance for both taxonomies were achieved with a plain transfer learning approach from the CORINE-pretrained model, without cropping augmentations or further semi-supervised learning. The overall performance of the baseline random forest model is better than any of the ResNet models alone, but ensembling these models with the random forest model boosts performance significantly. Altough the CORINE-classes are quite different from the final classes, pretraining a CNN with a large CORINE-dataset improves classification performance by a lot. 

It can be seen that both crop augmentation and semi-supervised learning lead to a worse performing model. The differences in macro-averaged F1 score are larger due to the nature of macro-averaging, where the score is biased towards smaller classes with few examples. Contrary to the original hypothesis, that semi-supervised learning and cropping would lead to the best performance, the best performing model applies neither of these. Layer freezing and pretraining however improve the model, as the model does not overfit the convolutional layers to the small dataset, and learns better representations from the CORINE-dataset.

\subsection{Qualitative evaluation}

Quantitative evaluation and performance metrics provide a numeric way for model performance assessment, but they fail to illustrate some differences between models. Qualitative evaluation, or the analysis of final classification maps and their differences provides an alternative approach to model evaluation. Some differences, such as classification uniformity, noise and granularity can only be seen from these maps and visual assessment.

Figure \ref{fig:score_suot} shows the classification confidence heatmaps for the Natura2000 class \textit{"7310 - Aapa mires"}. Typical to CNN-models, the ResNet model displays highly confident classifications for certain areas, while the Random Forest model is less confident and more fragmented. As the \ac{CNN} performs better with more mosaiced classes, such as wetlands, it is able to classify areas as aapa mires more confidently. The \ac{CNN} classifications are also more uniform, compared to the \ac{RF} predictions.

In general the ResNet detects larger and more mosaicked areas, such as wetlands, better, while losing some accuracy in finer-grained areas with rapid land-cover changes. Some rarer classes are found only in small areas and may not be detected by the ResNet model. The windowed nature of the ResNet classifier leads to misclassifications in transitional areas, like in rivers and surrounding wetlands. Often, when the center pixel to be classified hits a river, it is classified into a class surrounding the river, as seen around in Figure \ref{fig:class_river}.


\section{Conclusions}
\label{sec:conclusions}

We have discussed the challenging task of fine-grained habitat classification from remote sensed imagery, and proposed approaches for utilizing sparse pixelwise annotations with large amounts of unannotated data. We showed that pretraining a model with out-of-domain data (such as CORINE land cover labels) improves performance with fine-grained habitat classes. We also proposed a new model for single-pixel classification, using a ensemble of random forest classifier and a convolutional neural network, showing that the approach performs better than either model by themselves.

\bibliographystyle{elsarticle-num} 
\bibliography{references.bib}
\end{document}